\documentclass{article}

\usepackage{microtype}
\usepackage{graphicx}
\usepackage{subfigure}
\usepackage{booktabs} % for professional tables

\usepackage{hyperref}

\usepackage[accepted]{icml2025}

\usepackage{amsmath}
\usepackage{amssymb}
\usepackage{mathtools}
\usepackage{amsthm}
\usepackage{multirow}

\usepackage[capitalize,noabbrev]{cleveref}

\usepackage{lipsum}                     % Dummytext
\usepackage{xargs}                      % Use more than one optional parameter in a new commands
\usepackage[pdftex,dvipsnames]{xcolor}  % Coloured text etc.
\usepackage{etoolbox}
\usepackage{enumitem}

\usepackage[colorinlistoftodos,prependcaption,textsize=tiny]{todonotes}
\newcommandx{\unsure}[2][1=]{\todo[linecolor=red,backgroundcolor=red!25,bordercolor=red,#1]{#2}}
\newcommandx{\change}[2][1=]{\todo[linecolor=blue,backgroundcolor=blue!25,bordercolor=blue,#1]{#2}}
\newcommandx{\info}[2][1=]{\todo[linecolor=OliveGreen,backgroundcolor=OliveGreen!25,bordercolor=OliveGreen,#1]{#2}}
\newcommandx{\improvement}[2][1=]{\todo[linecolor=Plum,backgroundcolor=Plum!25,bordercolor=Plum,#1]{#2}}
\newcommandx{\thiswillnotshow}[2][1=]{\todo[disable,#1]{#2}}

\theoremstyle{plain}

\theoremstyle{definition}

\theoremstyle{remark}

\usepackage[textsize=tiny]{todonotes}

\newcounter{requirement}
\icmltitlerunning{Reusing Overtrained Language Models Saturates Scaling}

\begin{document}

\twocolumn[
\icmltitle{Reusing Overtrained Language Models Saturates Scaling}



\icmlsetsymbol{equal}{*}

\begin{icmlauthorlist}
% \icmlauthor{Firstname1 Lastname1}{equal,yyy}
% \icmlauthor{Firstname2 Lastname2}{equal,yyy,comp}
\icmlauthor{Seng Pei Liew}{comp}
\icmlauthor{Takuya Kato}{comp}
\end{icmlauthorlist}

\icmlaffiliation{comp}{SB Intuitions, Tokyo, Japan}

\icmlcorrespondingauthor{Seng Pei Liew}{sengpei.liew@sbintuitions.co.jp}

\icmlkeywords{language modeling, scaling law}

\vskip 0.3in
]



\printAffiliationsAndNotice{}  % leave blank if no need to mention equal contribution




\begin{abstract}
Reusing pretrained base models for further pretraining, such as continual pretraining or model growth, is promising at reducing the cost of training language models from scratch.
However, the effectiveness remains unclear, especially when applied to overtrained base models.
In this work, we empirically study the scaling properties of model reuse and find that the scaling efficiency diminishes in a predictable manner: The scaling exponent with respect to second-stage training tokens decreases logarithmically with the number of tokens used to pretrain the base model.
 The joint dependence on first- and second-stage tokens is accurately modeled by a simple scaling law.
 Such saturation effect reveals a fundamental trade-off in multi-stage pretraining strategies: the more extensively a base model is pretrained, the less benefit additional pretraining provides.
 Our findings provide practical insights for efficient language model training and raise important considerations for the reuse of overtrained models.
\end{abstract}

\section{Introduction}
\label{sec:introduction}
Large language models (LLMs) have recently shown astounding performance in various natural language processing tasks, reaching human-level capabilities in some cases \citep{achiam2023gpt,claude3,geminiteam2023gemini}. 
However, training/pretraining these models from scratch is computationally expensive and time-consuming, requiring weeks to months even with powerful GPU clusters. 
To address this challenge \emph{within the pretraining stage itself}, researchers have explored strategies for reusing existing pretrained (base) models for various purposes.
These include strategies to learn with new pretraining data, or to increase the model size, without starting from scratch.
Such strategies, which pretrain a model through multiple stages, are collectively referred to as \emph{model reuse} throughout this paper.

More specifically, we are interested in strategies that (1) improve domain-specific performance through \emph{continual pretraining} (CPT) of the base model (see, e.g., \citet{ibrahim2024simple}); or (2) scale model capacity via \emph{model growth} techniques, which increase the model size reusing base model parameters to speed up training \citep{chen2015net2net}.
These strategies have been shown to be promising at accelerating performance improvement of language models while reducing computational costs of pretraining from scratch.

On the other hand, {when a base model is overtrained, intuitively, its highly optimized parameters make effective exploration after capacity enlargement for model growth (or exploration with new data distribution for CPT) difficult, causing the second stage to learn slower.}
This is closely related to the loss of plasticity for neural networks over changing tasks \citep{ash2020warm,sodhani2020toward,lyle2023understanding,lyle2024disentangling}.
% This is known as the loss of plasticity for neural networks over changing   \citep{berariu2021study,ash2020warm,sodhani2020toward}.
% Grown models may also similarly suffer from bad initialization, resulting in ineffective utilization of their enlarged capacity.
% , and thus may not be able to learn new tasks effectively.
% are known to lose their plasticity, or ability to learn new tasks, as they are trained for longer periods of time.
Hence, it is unclear if the second stage of pretraining can be effectively applied to language models that are overtrained.

\begin{figure*}[ht!]
    \centering
        \includegraphics[width=0.3\linewidth]{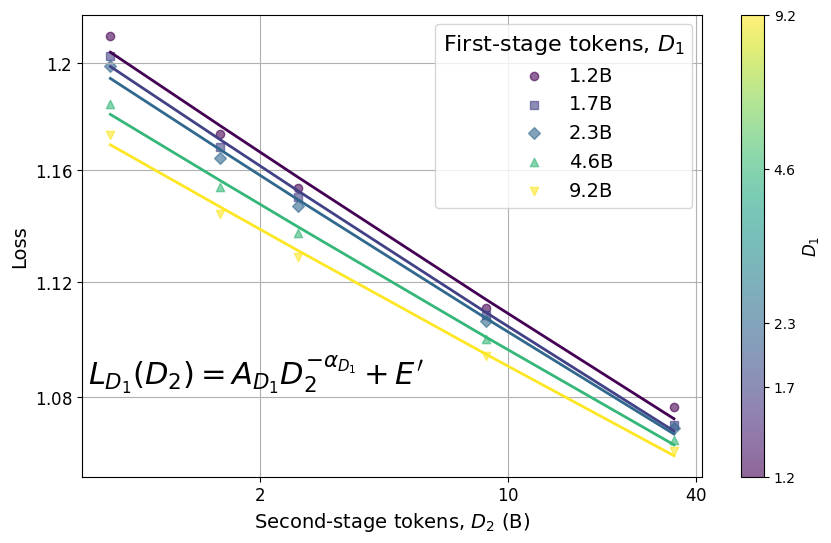}
        \includegraphics[width=0.3\linewidth]{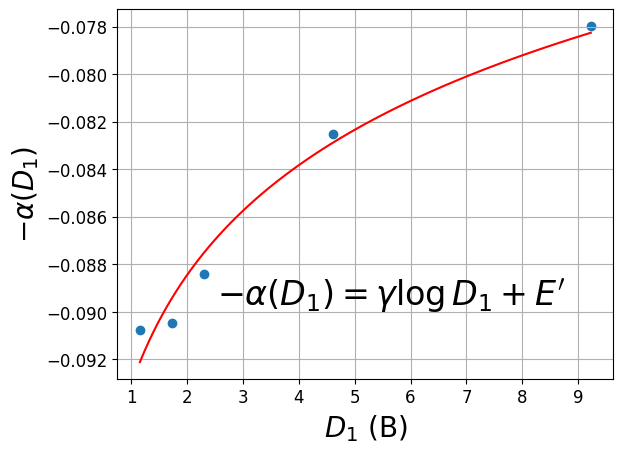}
        \includegraphics[width=0.3\linewidth]{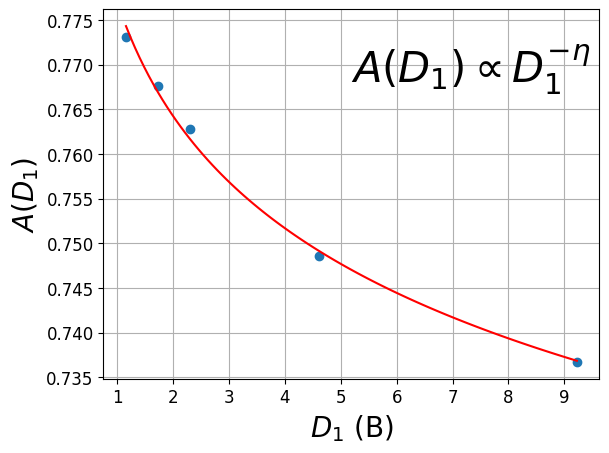}
        \includegraphics[width=0.3\linewidth]{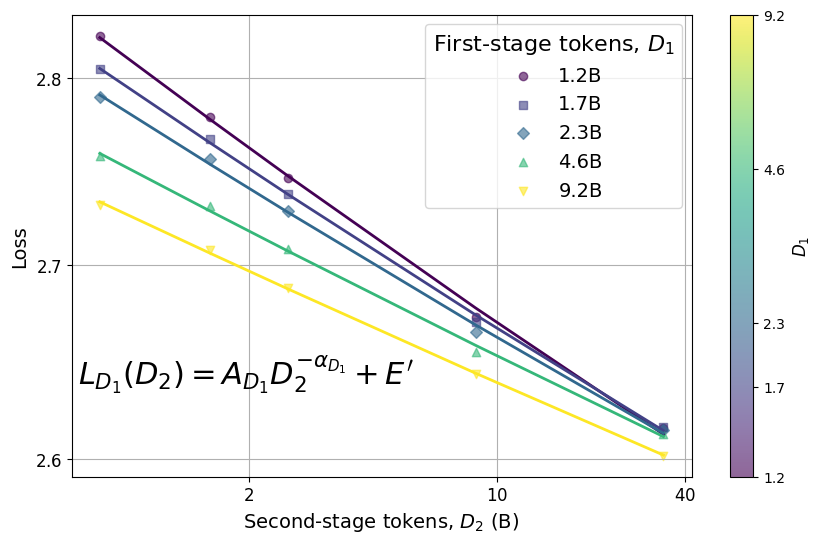}
        \includegraphics[width=0.3\linewidth]{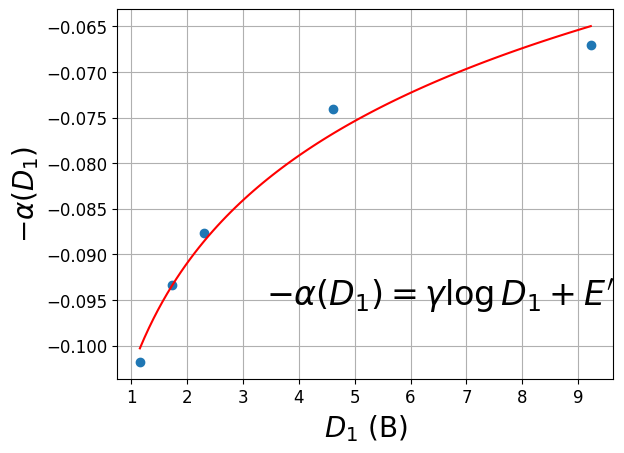}
        \includegraphics[width=0.3\linewidth]{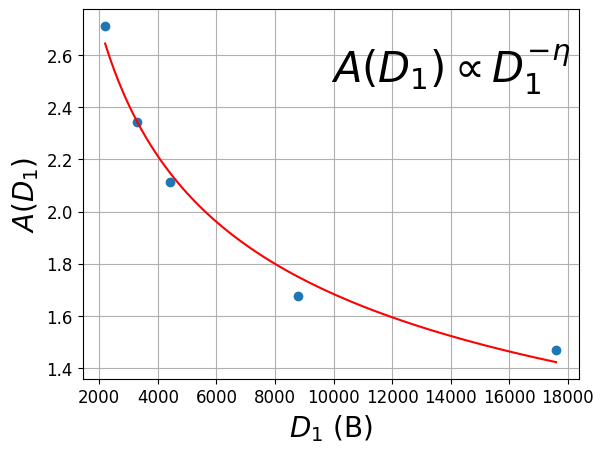}
    \caption{
              \textbf{Model reuse with overtrained base models leads to saturation in scaling behavior.} 
        \textbf{Left: $D_2$ has power-law scaling.} 
        We show scaling behavior of second-stage training tokens ($D_2$) for different values of first-stage tokens ($D_1$), trained on a 0.1B base model.
        \textbf{Middle: Interaction term explains decreasing exponents.} The fitted exponents in the left plots are used to fit Equation \ref{eq:log} as a function of $D_1$, and are shown to agree well with the functional form. 
               \textbf{Right: Scaling factor has power-law scaling w.r.t. $D_1$.}
               \textbf{Top: Continual pretraining (CPT) on code data. Bottom: Model growth from 0.1B to 0.2B by stacking.}  
        }    \label{fig:dataonly}
\end{figure*}

In this paper, we seek to understand the scaling properties of such methods by running a multitude of controlled language modeling experiments.
We find that, from the performance (cross-entropy loss) perspective, reusing overtrained base models leads to \emph{saturation} in scaling during the second-stage pretraining.
Specifically, we show that this effect is \emph{quantifiable} via scaling laws: 
\textbf{The scaling exponent (with respect to the number of training tokens in the second stage) decreases as the base model is trained for longer periods of time.
% Empirically, we show that these methods follow a multiplicative scaling law with respect to the number of training tokens invested to the first and second stage, and the scaling exponents of these methods ``decay" as the base model is trained for longer periods of time.
The decrease is proportional to the logarithm of the number of training tokens invested to the first stage.}
% This is particularly pronounced in the case of model growth.
See Figure \ref{fig:dataonly}.
Our work bears practical implications as even relatively tiny models are overtrained to trillions of tokens nowadays \citep{zhang2024tinyllama,dubey2024llama}, and model reuse strategies are widely adopted.

\textbf{Summary of contributions.}
More technically, our key contributions are as follows:
\begin{itemize}[leftmargin=*]
   %  \item We formulate possible functional forms for the scaling laws of model reuse methods, and empirically validate them on two representative methods: continual pretraining and model growth.
    \item We conduct extensive experiments on language models of various sizes and pretraining tokens/datasets (with over 450 runs), to study the scaling properties of model reuse methods, including continual pretraining and model growth.
    \item 
    We find that for a wide range of configuration of these methods, the following empirical scaling relation holds: 
    \begin{align}
    L(D_1,D_2) = A D_1^{-\alpha_1} D_2^{-\alpha_2 + \alpha_3 \log D_1} + E
    \label{eq:major}
\end{align}
    where $L$ is the validation loss after the second stage, $D_1$ and $D_2$ are the number of training tokens in the first and second stages, respectively, and $A,\alpha_1,\alpha_2,\alpha_3,E$ are \emph{positive} constants.
    We denote the term $\alpha_3 \log D_1$ as the \emph{interaction term}.
    The scaling exponent with respect to $D_2$ then quantifies the saturation effect of model reuse.
{We also formulate other plausible functional forms to make comparisons, showing that the above formula still best fits the empirical observation.}
            \item We further analyze the scaling properties with respect to model size, and show that the scaling laws can be extended to jointly incorporate model size together with dataset sizes, see Equation \ref{eq:joint}.  
        Limitations of our scaling law formulation when training with an extremely large number of tokens are also discussed.
        \item {We demonstrate how the scaling laws can provide guidance on when to train from scratch instead of model reuse to mitigate the negative effects of overtraining. 
        Moreover, we analyze the saturation phenomenon both mechanistically and via a toy model.}
\end{itemize}
Notations used throughout the paper are summarized in Appendix \ref{app:notation}.

% \textbf{Paper organization.}
% We first provide preliminaries in Section \ref{sec:prem}.
% Then, we present experiments for probing the scaling behavior in Section \ref{sec:exp}.
% The scaling laws are derived in Section \ref{sec:law}, followed by their implications for practitioners in Section \ref{sec:application}.  
% We discuss related work in Section \ref{sec:related} before closing with discussions in Section \ref{sec:conclusion}.

\subsection{Related Work}
\label{subsec:related}
\textbf{Continual pretraining.}
 CPT has been widely used to adapt existing LLMs to specific domains \citep{sun2020understanding,jang2022temporal,jang2022continual}, including code \citep{yadav2023continualcode,zan2022cert,guo2024deepseek} and mathematics \citep{gong2022continualpretraining,shao2024deepseekmath} domains, to be studied in this work.
Systematic studies at scale are relatively fewer, and our methodology mainly follows \citet{ibrahim2024simple}. 

\textbf{Model growth.}
While \citet{chen2015net2net} was the first model growth work in the deep learning era, the idea of growing neural networks from smaller ones can be traced to the 90s \citep{NIPS1989_69adc1e1,Fahlman1990TheRC}.
More recent language model-related model growth methods include \citet{gong2019efficient,chen2021bert2bert,wang2023learning,wang2023lemon,shen2022staged, evci2022gradmax,yao2024masked}, which were subsequently systematically analyzed in \citet{du2024stacking}, motivating our choice of model growth methods.

\textbf{Power-law ansatz.}  
We assume that the final validation loss $L$ follows the widely observed \emph{power-law scaling} with respect to a single variable of interest, such as the number of training tokens or the model size~\citep{hestness2017deep,hestness2019beyond,henighan2020scaling}:
 \begin{equation}
    L = A X^{-\alpha} + E, 
\label{eq:ansatz}
 \end{equation}
 where $X$ is the variable of interest, $\alpha$ is the scaling exponent, $A$ is the scaling factor, and $E$ is the irreducible loss due to the inherent entropy of the data distribution.
\footnote{ \label{fn:power}
Strictly speaking, we assume the form 
$L = \tfrac{A}{(X + 1)^\alpha} + E$, 
which ensures finiteness at $X \to 0$.  
However, since in practice $X \gg 1$ (typically $10^6$ or more), we approximate it as 
$L \approx AX^{-\alpha} + E$ for notational simplicity.
Moreover, we use the same symbols $A$, $\alpha$, and $E$ to denote the scaling factor, exponent, and additive constant, respectively, for notational conveniences when there is no ambiguity, although they may differ across different variables of interest.
}

% \textbf{Scaling laws.}
% We focus on modeling only the \emph{final} loss, which has a power-law like behavior as discussed above, a generic phenomenon not only occuring in neural networks, but is also observed in other natural as well as man-made phenomena \citep{clauset2009power}.
% We shall additionally note that recent scaling law studies attempted to fit the whole loss curve albeit with more sophisticated functional forms \citet{tissue2024scaling,wang2025learning,qiu2025scaling}.

% Moreover, we do not consider the scaling behavior with respect to other method-specific factors, such as the CPT replay ratio \citet{que2024d,wang2025learning} and the expansion factor in model growth \citet{du2024stacking}, as our focus is on the more general method-agnostic scaling behavior with respect to training tokens.

Similar to CPT, the reduced capability of transfer learning of code data from models pretrained on natural language data was observed in \citet{hernandez2021scaling}, where the authors denoted as ossification.
A similar phenomenon was also observed in the pretraining-fine-tuning pipeline \citep{springer2025overtrained}.
These studies however did not quantify the saturated scaling behavior.
Previous fine-tuning scaling studies \citep{mikami2022scaling,zhang2024scaling,bethune2025scaling} did not observe the subtler saturation effects as well, which is perhaps due to the smaller scale (in terms of training tokens) of their experiments.

\section{Experiments}
\label{sec:exp}
\subsection{Setup}
% In this section, we describe our experimental setup, including the model architectures, datasets, model reuse strategies, and training configurations.

\textbf{Model configuration.}
We consider decoder-only transformers~\citep{vaswani2017attention} pretrained with an autoregressive language modeling objective.  
Our architecture follows LLaMA~\citep{touvron2023llama}, incorporating refinements such as SwiGLU activation functions~\citep{shazeer2020glu} and rotary position embeddings~\citep{su2024roformer}.  
We use the LLaMA tokenizer with a vocabulary size of 32,000.
% We consider a suite of model sizes up to 1.1B in our experiments.
Table \ref{tab:config} of Appendix \ref{app:exp} contains the model configuration used in this paper.
% Table \ref{tab:config} of Appendix \ref{app:exp} contains the model configuration (number of layers, $n_{\rm layer}$, hidden dimension $d_{\rm model}$, and  MLP hidden dimension, $d_{\rm mlp}$) used in this paper.
% Table~\ref{tab:config} in Appendix contains the main architectural configuration of these models.

\textbf{Training configuration.}
All experiments are conducted using the Megatron-LM library~\citep{shoeybi2019megatron}.  
We train models in mixed precision (bfloat16) using the AdamW optimizer~\citep{loshchilov2017fixing}, with maximum learning rates tuned separately for each model size.
The learning rate follows the \emph{warmup-stable-decay} (WSD) schedule~\citep{bi2024deepseek,hu2024minicpm}, with the final learning rate decayed to one tenth of the peak value.
This schedule enables long training runs and facilitates checkpoint reuse for emulating shorter effective training budgets, reducing computational cost.
In Appendix \ref{app:train}, we also provide an ablation study with the cosine learning rate schedule to show that they achieve similar performance.

We train the models for a number of tokens to at least roughly 200 times the model size in tokens in total (token-to-parameter ratio).
Tables \ref{tab:common} and \ref{tab:train_config} in Appendix summarize the main training configurations.
See also Appendix~\ref{app:exp} for further experimental details and Appendix \ref{app:train} for training details and evaluation on downstream tasks.

\begin{figure}{}
% \begin{figure}
   \centering 
    \includegraphics[width=0.5\textwidth]{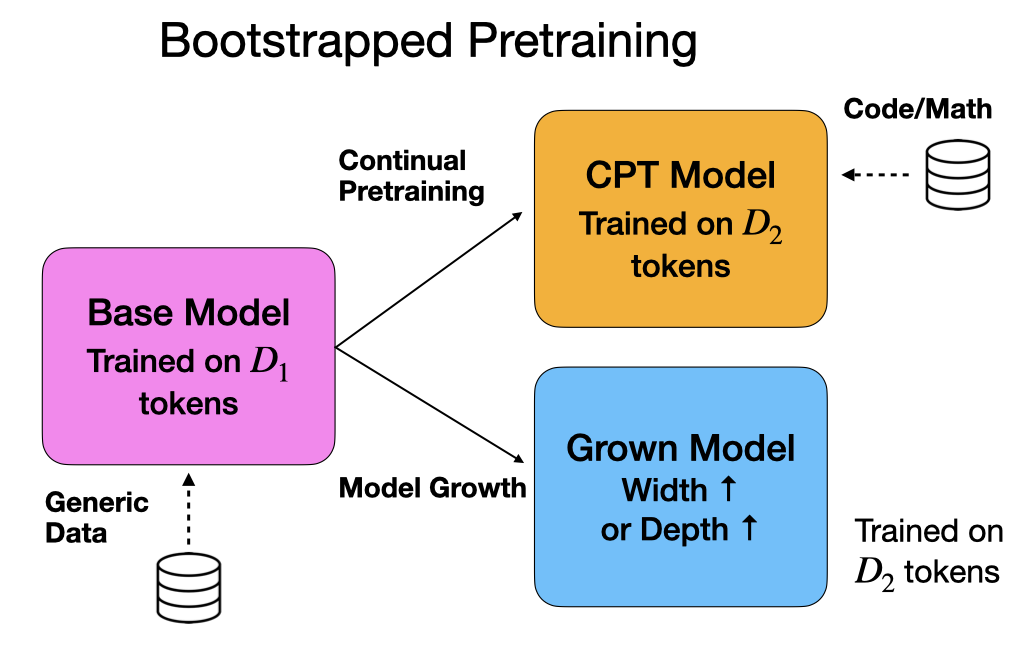}
    \caption{\textbf{Illustration of model reuse in consideration.} model reuse consists of two stages: (1) first-stage pretraining of a base model for $D_1$ tokens on internet/generic data; (2) second-stage pretraining via continual pretraining or model growth for $D_2$ tokens. Section \ref{sec:formulation} and \ref{sec:data} study and develop scaling laws as a function of these two variables (and additionally model size, $N$ in Section \ref{sec:closer}) to predict the final loss after the second stage.}
    \label{fig:method}
\end{figure} 
\textbf{Two stages of pretraining.}  
In the context of model reuse, we are primarily interested in how the loss behaves with respect to the number of training tokens used in the two stages of pretraining:
\begin{itemize}[leftmargin=*]
    \item $D_1$: the number of tokens used in the \textbf{first-stage} pretraining (base model).
    \item $D_2$: the number of tokens used in the \textbf{second-stage} pretraining (CPT or model growth).
\end{itemize}

\textbf{Base model.}
For the first-stage pretraining, we use the CommonCrawl portion of the Slimpajama-DC dataset~\citep{shen2023slimpajama}, containing 368B tokens in total.  
The models trained in this stage serve as base checkpoints for the second-stage model reuse experiments.

\textbf{Continual Pretraining.}
CPT refers to further training a pretrained model on a large dataset, typically billions of tokens, with the goal of improving performance on a new domain.  
We distinguish CPT from \emph{instruction tuning}, which typically trains with smaller datasets (millions of tokens or fewer), and is thus harder to analyze the scaling behavior.
For our experiments, we perform CPT on the base model with two domain-specific datasets: \textbf{Code corpus:} Stack/StarCoder~\citep{li2023starcoder}; and \textbf{mathematics corpus:} OpenWebMath~\citep{paster2023openwebmath}.
% \begin{itemize}
%     \item \textbf{Code corpus:} Stack/StarCoder~\citet{li2023starcoder}.
%     \item \textbf{Mathematics corpus:} OpenWebMath~\citet{paster2023openwebmath}.
% \end{itemize}
Unless otherwise noted, we adopt the same optimizer and learning rate configurations as the first-stage pretraining.

\textbf{Model growth.}
Model growth refers to increasing model capacity by adding new layers and/or expanding hidden dimensions, thereby increasing the number of trainable parameters.  
This strategy aims to leverage the representations learned during first-stage training, allowing the larger model to accelerate learning in the second stage.
We investigate two model growth techniques shown to be the most effective (in terms of adding new parameters in the width and depth directions) in prior work~\citep{du2024stacking} (see also Appendix \ref{app:mg} for more details):

\begin{itemize}[leftmargin=*]
    \item \textbf{Width expansion (exp):} Add new neurons to each layer while using \emph{function-preserving initialization} (FPI), ensuring that the expanded model initially reproduces the behavior of the smaller model.
    \item \textbf{Depth-wise stacking (stk):} Add new layers to the top of the existing one by direct copying, thereby extending model depth.
\end{itemize}

After performing this procedure, the larger models are trained on the same dataset as the base model. 
The growth factor is defined as the ratio of the number of non-embedding parameters in the grown model to that in the base model.
For both model reuse scenarios in consideration, the validation loss is evaluated on a held-out set from the second-stage dataset.
We illustrate the whole framework in Figure~\ref{fig:method}.

% \end{figure}
% to isolate the effect of growth from data distribution shifts.

% , which consists of three phases:
% \begin{enumerate}
%     \item Linear warmup to peak learning rate.
%     \item A stable training phase with constant learning rate.
%     \item Linear decay to zero.
% \end{enumerate}

\subsection{Experimental Observations}
{We here provide experimental observations that motivate our formulation of the scaling law.}
In Figure \ref{fig:dataonly}, we show results of training a base model of size 0.1B, and performing CPT on code data, and stacking with a growth factor of 2 on a $5\times5$ grid of $D_1,D_2$. \footnote{We double the number of non-embedding layers for stacking.} 
In the left panels, we plot the second-stage validation loss after the second stage as a function of $D_2$ for different values of $D_1$ (left).
The following observation is made:
% we use a model of size 0.1B, trained on a $5\times5$ grid of $D_1,D_2$. 
% \textbf{Power-law ansatz.}  
% We begin by assuming that the validation loss $L$ follows the widely observed \emph{power-law scaling} with respect to a single variable of interest, such as the number of training tokens or the model size~\citep{hestness2017deep,hestness2019beyond,henighan2020scaling}: $$L = A X^{-\alpha} + E,$$ where $X$ is the variable of interest, $\alpha$ is the scaling exponent, $A$ is the scaling factor, and $E$ is the irreducible loss due to the inherent entropy of the data distribution.
% \footnote{ \label{fn:power}
% Strictly speaking, we assume the form 
% $L = \tfrac{A}{(X + 1)^\alpha} + E$, 
% which ensures finiteness at $X \to 0$.  
% However, since in practice $X \gg 1$ (typically $10^6$ or more), we approximate it as 
% $L \approx AX^{-\alpha} + E$ for notational simplicity.
% Moreover, we use the same symbols $A$, $\alpha$, and $E$ to denote the scaling factor, exponent, and additive constant, respectively, for notational conveniences, although they may differ across different variables of interest.
% }

{
\textbf{Observation 1:}  
The loss is observed to follow a power law with respect to the number of tokens $D_2$, i.e., Equation \ref{eq:ansatz}.}
% \begin{equation}
%     L(D_1,D_2) = L_{D_1}(D_2) = A D_2^{-\alpha} + E.
%     \label{eq:power_d2}
% \end{equation}

This is consistent with previous scaling laws reflecting the expectation that, when starting from a fixed initialization, additional tokens improve performance predictably.

Furthermore, we observe that the fitted scaling exponent of Equation \ref{eq:ansatz} decreases as $D_1$ increases, with the model growth method more pronouncedly so.
% To quantify this relationship, we model the scaling exponent as a function of $D_1$, denoted as $-\alpha(D_1)$. A scatter plot of $-\alpha(D_1)$ reveals the following logarithmic relationship:
To quantify this trend, we express (the minus of) the scaling exponent as a function of $D_1$, denoted by $-\alpha(D_1)$.
A scatter plot of $-\alpha(D_1)$ reveals a clear logarithmic dependence:
\begin{equation}
-\alpha(D_1) = \gamma \log D_1 + E'. \label{eq:log}
\end{equation}
% which fits the data well, as shown in the same Figure. 
This relationship fits the empirical data well, as shown in the same Figure.
% Substituting this expression to Equation \ref*{eq:power_d2}, we obtain a term of the form $ D_2^{- E'+\gamma \log D_1}$, directly supporting the interaction term in the multiplicative scaling law. 
Substituting Equation~\ref{eq:log} into Equation~\ref{eq:ansatz}, we arrive at a term of the form $D_2^{-E' + \gamma \log D_1}$.
Additionally in the same Figure, we show that the multiplicative dependence, $A\propto D_1^{-\alpha_1}$, also holds well.
These lead to the following observation:

\textbf{Observation 2:}  
The dependencies of the second-stage loss jointly on $D_1$ and $D_2$ can be captured by a multiplicative scaling law with an interaction term, i.e., Equation \ref{eq:major}.

\textbf{Interpreting the scaling law.}
Several insights can be made from the scaling law.
Fixing $D_1$, the effective scaling factor for $D_2$ becomes $A D_1^{-\alpha_1}$.
As $D_1$ increases, this factor decreases, resulting in a lower initial loss at the start of second-stage pretraining, agreeing with the conventional wisdom that better-pretrained base models (larger $D_1$) provide stronger initialization for second-stage pretraining.

However, also at fixed $D_1$, the effective scaling exponent with respect to $D_2$ is given by $\alpha_2 - \alpha_3 \log D_1$.
This implies that as the base model becomes more overtrained (i.e., larger $D_1$), the improvement in loss from additional second-stage tokens becomes increasingly marginal.
In other words, the returns from increasing $D_2$ diminish, manifesting as saturation effects at higher $D_1$.

\section{Scaling Laws for Model Reuse}
\label{sec:formulation}
\subsection{Functional Forms for the Scaling Laws}
% Observations from the previous Section on the functional form of the scaling law are purely empirical.
% One may wonder if there exist other possible functional forms that could fit the observation.

The observations from the previous Section regarding the functional form of the scaling law are purely empirical, which naturally raises the question of whether other functional forms might also fit the observed data.
To explore these possibilities, we first derive functional forms that satisfy certain reasonable assumptions, and later fit and compare them empirically with our experimental results.
% \textbf{Power-law ansatz.}  
% We begin by assuming that the validation loss $L$ follows the widely observed \emph{power-law scaling} with respect to a single variable of interest, such as the number of training tokens or the model size~\citep{hestness2017deep,hestness2019beyond,henighan2020scaling}: $$L = A X^{-\alpha} + E,$$ where $X$ is the variable of interest, $\alpha$ is the scaling exponent, $A$ is the scaling factor, and $E$ is the irreducible loss due to the inherent entropy of the data distribution.
% \footnote{ \label{fn:power}
% Strictly speaking, we assume the form 
% $L = \tfrac{A}{(X + 1)^\alpha} + E$, 
% which ensures finiteness at $X \to 0$.  
% However, since in practice $X \gg 1$ (typically $10^6$ or more), we approximate it as 
% $L \approx AX^{-\alpha} + E$ for notational simplicity.
% Moreover, we use the same symbols $A$, $\alpha$, and $E$ to denote the scaling factor, exponent, and additive constant, respectively, for notational conveniences, although they may differ across different variables of interest.
% }

% In the context of model reuse, we are interested in modeling the loss as a function of \emph{two} variables: $D_1, D_2$.
To recite, our goal is to find a functional form $L(D_1, D_2)$ that captures how the second-stage loss depends jointly on both stages.
% \textbf{Conditions for the scaling law.}  
To derive a principled formulation, we impose two natural constraints:

\textbf{Condition 1:}  
For a fixed base model trained on $D_1$ tokens, the second-stage loss should follow a power law with respect to the number of tokens $D_2$:
\begin{equation}
    L(D_1,D_2) = L_{D_1}(D_2) = A_{D_1} D_2^{-\alpha_{D_1}} + E_{D_1}.
    \label{eq:power_d2}
\end{equation}
This is consistent with classical neural scaling laws and Observation 1.
 It reflects the expectation that, when starting from a fixed initialization, additional tokens improve performance predictably.

\textbf{Condition 2:}  
For any fixed value of $D_2$, the loss should exhibit power-law behavior with respect to the number of first-stage tokens $D_1$:
\begin{equation}
L(D_1, D_2) = L_{D_2}(D_1) = A_{D_2} D_1^{-\alpha_{D_2}} + E_{D_2}
\end{equation}
This is consistent with the power-law ansatz and captures the intuition that a better-trained base model (i.e., larger $D_1$) should result in lower loss.

Moreover, this condition implies that as $D_2 \to 0$, the loss should continuously approach that of the base model: $\lim_{D_2 \to 0} L(D_1,D_2) = A D_1^{-\alpha} + E$ (as in Footnote \ref{fn:power}, we omit the $+1$ term in $D+1$ expressions for notational simplicity, such that terms proportional to $D_2$ are finite as $D_2 \to 0$), which is a natural requirement for \textit{function-preserving model growth}; when model capacity is expanded but initialized carefully, the initial loss should remain close to the base model’s loss.
% \begin{itemize}[leftmargin=*]
%     \item \textbf{CPT:} The second stage's initial loss should begin from the base model’s (evaluated on second-stage dataset).
%     \item \textbf{Function-preserving model growth:} When model capacity is expanded but initialized carefully, the initial loss should remain close to the base model’s loss.
%     %  \footnote{
% % For stacking-based model growth, the function-preserving property does not hold and the initial loss may be higher than the base model.
% % We nevertheless retain this assumption for completeness.
% % }
% \end{itemize}
% \textbf{Functional forms.}  
We further validate this Condition by showing empirically that the loss follows a power law with respect to $D_1$ for fixed $D_2$ in Appendix \ref{app:scaling}.
% We further show that the loss follows a power law with respect to $D_1$ for fixed $D_2$ in Figure \ref{fig:d1_dependence} (with more plots in Appendix \ref{app:scaling}).
% This validates Condition 2 we impose for deriving the scaling laws.
% Note that Condition 1 is directly implied by Observation 1.

Together, these conditions lead to the following candidate formulations that jointly satisfy both requirements:
% \textbf{(1) Multiplicative form}
% \begin{align}
%   \textbf{Multiplicative:  }  L(D_1,D_2)
%     &=  A D_1^{-\alpha_1} D_2^{-\alpha_2 + \alpha_3 \log D_1} + E.
%     \label{eq:multiplicative}
% \end{align}
% This formulation captures an \emph{interaction effect} between $D_1$ and $D_2$ through the $\alpha_3 \log D_1$ term.
% Intuitively, as the base model becomes more trained ($D_1$ large), additional data $D_2$ yields \textbf{diminishing returns}, reflected in a smaller effective exponent.
\begin{align}
  \textbf{Multiplicative:} \;
  L(D_1, D_2)
    &= A D_1^{-\alpha_1} D_2^{-\alpha_2 + \alpha_3 \log D_1} + E.
    \label{eq:multiplicative}
\end{align}
% \textbf{(2) Additive form}
\begin{align}
  \textbf{Additive:} \;
    L(D_1,D_2)
    &=  A D_1^{-\alpha_1} + F D_2^{-\alpha_2} + E.
    \label{eq:additive}
\end{align}
% This assumes first- and second-stage contributions are \emph{independent}, making it suitable when the two stages operate in largely disjoint feature spaces (e.g., cross-domain CPT).
% \textbf{(3) Hybrid form}
\begin{align}
  \textbf{Hybrid:} \;
    L(D_1,D_2)
    &=  \big(A D_1^{-\alpha_1} + F\big) D_2^{-\alpha_2} + E.
    \label{eq:hybrid}
\end{align}
Note that since $D_1^{-\alpha_1}D_2^{-\alpha_2+\alpha_3\log D_1} = D_1^{-\alpha_2+\alpha_3\log D_2}D_2^{-\alpha_2}$, the multiplicative form with an interaction term satisfies our conditions, while other forms do not have such an interaction term.
% This formulation combines the advantages of the multiplicative and additive forms, making it suitable when the base model provides a strong initialization but the second stage still dominates learning dynamics.
We further note that multiplicative scaling laws have been observed in \citet{mikami2022scaling,zhang2024scaling} but without the interaction term.
\citet{pmlr-v267-liew25a} empirically showed that sparse upcycling (training sparse mixture-of-experts (MoE) models reusing existing dense models) follows a scaling law similar to Equation \ref{eq:multiplicative} but under different motivation and conditions.
Our work directly extends their findings to other pretraining paradigms.
Equation \ref{eq:hybrid} has also been studied in \citet{barnett2024empirical}.

\subsection{Fitting Scaling Laws}
% \liew{add more explanation of fitting}
\label{sec:data}
% \textbf{Fitting the scaling laws.}
{
To determine the functional form described in the previous Section that best describes the scaling behavior of model reuse methods, we consider a variety of datasets and methods beyond those described in the previous Section.}

\textbf{CPT.} We consider both code and mathematics datasets and perform CPT on them as in the previous Section.
Additionally, we consider the following variants of CPT that are commonly used in practice \citep{ibrahim2024simple} and test them on the code dataset:
\begin{itemize}[leftmargin=*]
    \item \textbf{CPT with replay (rep):} A portion of the first-stage data is mixed into the second-stage data, which is common for mitigating catastrophic forgetting. 
    We consider a replay ratio of 0.25, i.e., 25\% of the second-stage data is from the first stage.
    \item \textbf{CPT from stable phase (sta):} We consider CPT starting from a base model checkpoint in the stable phase of the WSD learning rate schedule.
    This is to avoid adverse effects from re-warming the learning rate from a decayed value.
\end{itemize}

\textbf{Model growth.}
In addition to depth-wise stacking described in the previous Section, we consider width expansion that double the size of the base model. 
For a growth factor of 2, we increase the size of the hidden dimension by $\sqrt{2}$ for width expansion.
We further consider stacking variants:
\begin{itemize}[leftmargin=*]
    \item \textbf{Larger growth factor (x4):} We consider a larger growth factor of 4 times (instead of 2).
    \item \textbf{Model growth from stable phase (sta):} Similar to CPT, we train the stacked model starting from a base model checkpoint in the stable phase.
    %  This also avoids re-warming the learning rate from a decayed value.
\end{itemize}

\textbf{Comparing various functional forms.}
Running the experiments as described above and obtaining the results, we fit the losses with the functional forms of Equations \ref{eq:multiplicative}, \ref{eq:additive}, and \ref{eq:hybrid}.
Following \citet{hoffmann2022training,besiroglu2024chinchilla}, we perform optimization using the Huber loss ($\delta=10^{-3}$) and the BFGS algorithm, to fit the logarithm of the loss via the LogSumExp trick applied to the RHS of functional forms.
The leave-one-out root mean square error (RMS) serves as the goodness-of-fit metric.

\begin{figure*}[h]
    \centering
        \includegraphics[width=0.3\linewidth]{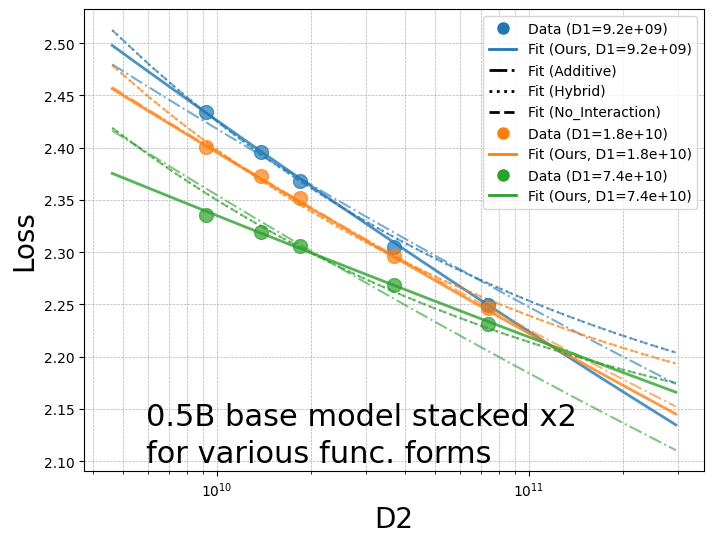}
            \includegraphics[width=0.3\textwidth]{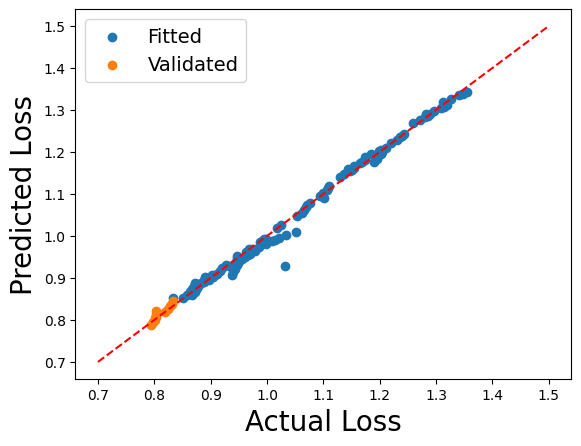}
            \includegraphics[width=0.3\textwidth]{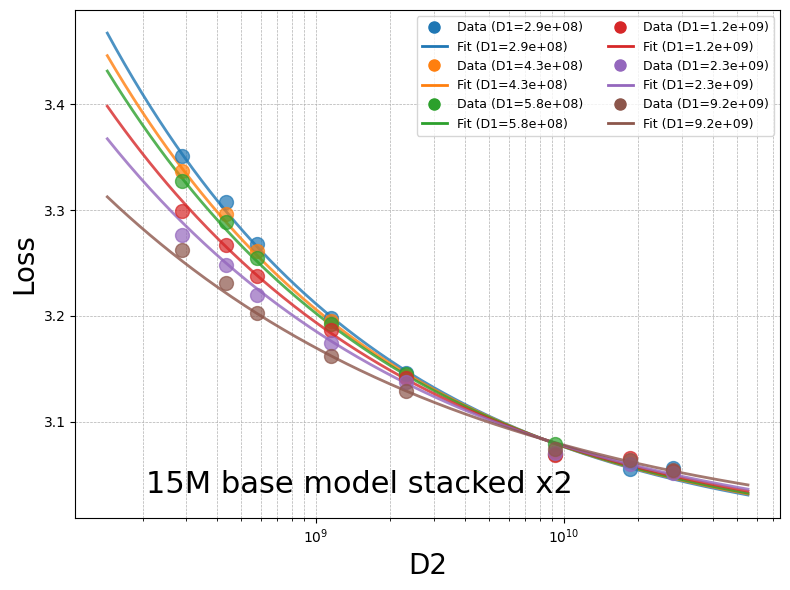}
    \caption{
        \textbf{Left: Visual comparison of various functional forms.} 
        We show scaling behavior of second-stage training tokens ($D_2$) for different values of first-stage tokens ($D_1$) for model growth by stacking (growth factor 2, 0.5B-to-1B model), comparing various functional forms. Our proposed multiplicative scaling law with interaction (solid lines) fits the data (dots) best, and lines with different $D_1$ cross over, indicating saturation effects. 
        Note that the hybrid functional form (dotted lines) nearly overlaps with the multiplicative without interaction form (dashed lines) in this plot.
        \textbf{Middle: Joint scaling law fit for continual pretraining on code data.} Orange points indicate the 10\% of data with lowest losses used for validation. \textbf{Right: Stacking with extremely large token-to-parameter ratio.} At the limit of large token ($D_2$)-to-parameter ratio, the values of the loss are too noisy to yield a definite characterization of the crossover region.}    \label{fig:d1_dependence}
        % \textbf{Left:} Continual pretraining on code data (1B model)
        % \textbf{Right: $D_1$ has power-law scaling.} 
        % We show scaling behavior of first-stage training tokens ($D_1$) for different values of second-stage tokens ($D_2$) for model growth by stacking (growth factor of 2), indicating that $D_1$ also has power-law scaling.
        %  More plots for other methods are available in Appendix \ref{app:scaling}.
\end{figure*}

As can be seen in Table \ref{tab:func_flipped}, the multiplicative form with interaction consistently achieves the lowest error across all methods and datasets, indicating that it best captures the underlying scaling behavior. 
To visualize these relationships, we compare the loss-versus-token plots of our proposed scaling law against alternative functional forms (Figure \ref{fig:d1_dependence}, left panel), using a 0.5B-to-1B model by stacking.
% Moreover, we visually compare loss-versus-token plots of our proposed scaling law against other functional forms in Figure \ref{fig:d1_dependence}, left panel.
The predicted losses from the alternative forms are visually confirmed to fit the empirical data worse.
% We see that the predicted losses of the alternative functional forms fit worse visually.
More importantly, the predicted losses from alternative forms with varying $D_1$ remain nearly parallel across the range of $D_2$, whereas ours converges, a distinct behavior that is not captured by other forms.
In Appendix \ref{app:cont}, we compare with yet another scaling formulation (``continuous" scaling law \cite{que2024d,tissue2024scaling}) and show that our scaling law still fits better.
% This demonstrates that beyond a certain training threshold, increased first-stage training ($D_{1}$) can even be \tex1tit{detrimental} to subsequent second-stage performance.
%  Furthermore, the predicted losses of other functional forms with different $D_1$ are nearly parallel across the range of $D_2$, whereas ours converge to a crossover point, and diverge, where losses with larger $D_1$ eventually become worse than those with smaller $D_1$.
% This means that beyond a certain point, more first-stage training can be detrimental to second-stage performance.

\begin{table*}
    \caption{
    \textbf{Multiplicative scaling law with interaction consistently achieves lowest error.} 
    Leave-one-out RMS error ($\times 10^{-3}$) for fitting the loss for model reuse of a 0.1B base model. Functional forms are from Equations \ref{eq:multiplicative} (and the specific case where $\alpha_3=0$), \ref{eq:additive}, and \ref{eq:hybrid}.
    }
    \label{tab:func_flipped}
    \centering
    \begin{tabular}{l|cccccccc}
        \toprule
        RMS $(\times 10^{-3})$ & \textbf{Code} & \textbf{Math} & \textbf{CPT (rep)} & \textbf{CPT (sta)} & \textbf{Exp x2} & \textbf{Stk x2} & \textbf{Stk x4} & \textbf{Stk (sta)} \\
        \midrule
        \textbf{Mul.} & \textbf{1.573} & \textbf{1.737} & \textbf{0.987} & \textbf{1.371} & \textbf{2.790} & \textbf{2.845} & \textbf{2.152} & \textbf{2.957} \\
        \textbf{Mul. ($\alpha_3=0$)} & 2.095 & 3.147 & 2.222 & 2.366 & 4.837 & 7.818 & 6.557 & 8.757 \\
        \textbf{Add.} & 3.913 & 7.443 & 4.646 & 4.315 & 9.855 & 10.323 & 8.866 & 10.559 \\
        \textbf{Hyb.} & 2.245 & 3.340 & 2.223 & 2.367 & 4.841 & 7.748 & 6.667 & 8.822 \\
        \bottomrule
    \end{tabular}
\end{table*}

Overall, the proposed scaling law is robust across datasets, configurations, and methods investigated.
While we acknowledge the possibility of edge cases that deviate from this law, it is important to note that our empirical validation specifically targeted training methodologies that are considered scalable and representative of current best practices in the field.
% we note that we validated the scaling law empirically on training methodologies considered scalable and representative of best practices.
 Consequently, we conclude that the scaling behavior and saturation effect identified here are not isolated phenomena, but are fundamentally representative of general model reuse methods employed in efficient LLM development.

\section{Extensions and Limitations}
\label{sec:closer}
 \subsection{Joint Scaling Incorporating Model Size}
We proceed to extend the data scaling law to include model size, $N$. 
Let us first focus on CPT where the model size remains unchanged after second-stage pretraining.
To determine the functional form, we impose that the loss follows the well-grounded ``Chinchilla" scaling law \citep{hoffmann2022training} (which jointly models the base model's loss with respect to dataset and model sizes) with respect to $D_2$ and $N$.

\textbf{Condition 3:}  
For a fixed base model trained on $D_1$ tokens, the second-stage loss should follow the Chinchilla scaling law with respect to the number of tokens $D_2$ and model size $N$:
$$L_{D_1}(D_2,N)= A_{D_1} D_2^{-\alpha_{D_1}} + N_{D_1}^{-\beta_{D_1}} + E_{D_1}.$$ 
% Note that this condition is consistent with the conditions (power-law ansatz) imposed in Section \ref{sec:formulation}.
The straightforward functional form that satisfies the Chinchilla-style scaling law is:
\begin{equation}
    \label{eq:joint}
    L(D_1,D_2,N)= A D_1^{-\alpha_1} D_2^{-\alpha_2 + \alpha_3 \log D_1} + BN^{-\beta} + E,
\end{equation}
which is also consistent with the conditions (power-law ansatz) imposed in Section \ref{sec:formulation}.
% which is also consistent with the multiplicative (with interaction) data scaling law obtained in Section \ref{sec:data} when $N$ is fixed.

\begin{figure*}
   \centering 
           \includegraphics[width=0.46\linewidth]{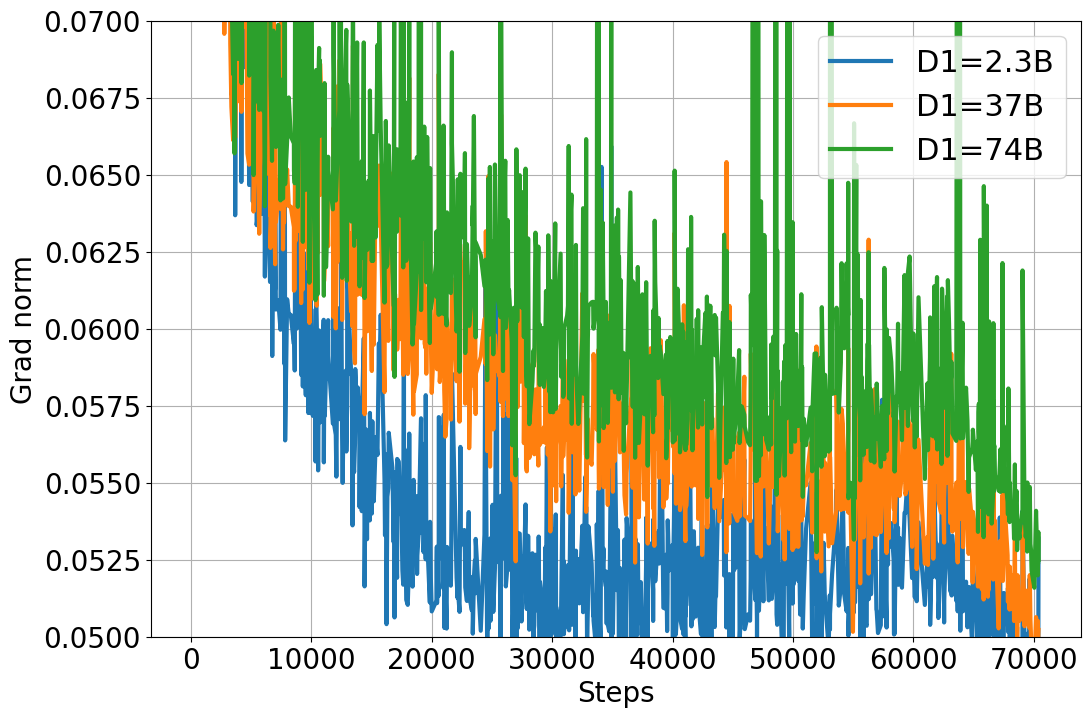}
       \includegraphics[width=0.42\textwidth]{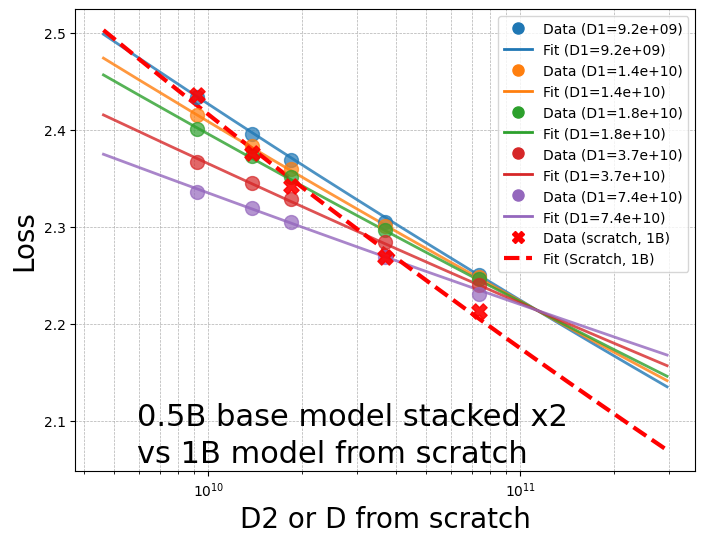}
    \caption{
    \textbf{Left: Overtrained models have larger gradient norms when undergoing model reuse.} 
        We show the gradient norm curves for continual pretraining on code data (1B model) with respect to number of training steps in the second stage for base models trained with different values of first-stage tokens ($D_1$). We see that overtrained models have larger gradient norms, indicating that the optimization difficulty increases with $D_1$, which may explain the saturation effect observed in the scaling law.
        See Appendix \ref{app:mechanistic} for model growth results, where similar trends are observed.    
    \textbf{Right: Loss versus second-stage (from-scratch) training token for stacking (training from scratch) a 0.5B-to-1B model (1B model)}. Data points and fitted lines using our multiplicative scaling law with different number of first-stage training tokens (power-law) are shown. It can be seen that as the number of training tokens increases, the losses from model growth saturate, and from-scratch training eventually outperforms it.}
    \label{fig:valid}
\end{figure*}

\textbf{Fitting the joint scaling law.}
We conduct experiments with base models of sizes 15 million (M), 44M, 0.1B, 0.2B, 0.5B and 1B, training them with different numbers of first and second-stage tokens as in Section \ref{sec:data}.
% Since we vary the base model sizes and train them with different $D_1$ values before performing the second stage pretraining, we can derive a separate joint scaling law for the base model as well.
To fit the joint scaling law, we use the same fitting procedure as in Section \ref{sec:data}, additionally taking $N$ as an additional variable. 
% \footnote{For model growth, we consider $N$ as the size of the model before growth, and fix the growth factor to 2 (i.e., $N$ doubles after growth).}
In the left panel of Figure \ref{fig:valid}, we show that the formula fits the data well for CPT on code.
%  and the goodness-of-fit to the validation data points also shows that it can be extrapolated to larger model and dataset sizes.
See Table \ref{tab:param} in Appendix \ref{app:joint} for the fitted coefficients. 
In the same Appendix, we further extend our results to model growth where model sizes increase. 

The joint scaling law allows us to extrapolate performance to models and datasets larger than those studied experimentally (Figure \ref{fig:d1_dependence}, middle panel).
We show in Appendix \ref{app:extrapolate} with a specific example how it can be used to predict the performance of larger models and datasets.
We can further study the compute optimality of CPT and model growth using the joint scaling law, as shown in Appendix \ref{app:compute}.
% Equipped with the joint scaling law, we can also study the compute optimality of model reuse, analogous to \citet{hoffmann2022training}. 
% See Appendix \ref{app:compute}.

% \begin{wrapfigure}{r}{0.5\textwidth}
\subsection{Limitations of the Scaling Law}
\label{sec:limit}
We dicusss scenarios where our scaling laws may not hold, especially when the number of training tokens is extremely large.

\textbf{Large $D_1$ limit.}
A theoretical limitation of Equation \ref{eq:major} is that as $D_{1}$ increases, the effective scaling exponent for $D_{2}$, $\alpha_{\text{eff}} = \alpha_{2}-\alpha_{3}\log~D_{1}$, decreases without bound, potentially becoming negative when $D_{1}>e^{\alpha_{2}/\alpha_{3}}$.
% Note that in Equation \ref{eq:major}, as $D_1$ increases, the effective scaling exponent $\alpha_2 - \alpha_3 \log D_1$ decreases without bound, and can even become negative when $D_1 > e^{\alpha_2/\alpha_3}$.
% However, we also note that $\alpha_2$ is often larger than $\alpha_3$ by at least an order of magnitude (see Table \ref{tab:param} in Appendix \ref{app:joint}), meaning that the threshold $e^{\alpha_2/\alpha_3}$ is often very large (e.g., $>10^{17}$).
% This is unlikely to be reached in practice.
However, based on our fitted coefficients where $\alpha_{2}$ often exceeds $\alpha_{3}$ by at least an order of magnitude (see Table \ref{tab:param} in Appendix \ref{app:joint}), this theoretical threshold (e.g., $D_1>10^{17}$) is unlikely to be reached in practice. 
We anticipate the scaling law would break down well before this point.

% Even if this is reachable, we expect that the scaling law would break down before reaching the threshold.
% where the effective exponent becomes negative, as this would imply that increasing $D_2$ would \emph{increase} the loss, which is counter-intuitive.
Alternatively, one can formally model this inevitable thresholding behavior, which marks a deviation from the core power-law ansatz.
We propose a modification motivated by \citet{clark2022unified} (used for MoE scaling laws):
% Alternatively, we can model the thresholding behavior which deviates from the power-law ansatz as follows, motivated by \citet{clark2022unified} which attempted to model the thresholding effects from the number of experts in the MoE scaling law.
In Equation \ref{eq:major}, we replace $D_1$ with a saturated form, $\hat{D_1}$, defined as
\begin{equation}
    \hat{D_1} = \left(D_1^{-1} + D_{\rm max}^{-1} \right)^{-1},
\end{equation}
where $D_{\rm max}$ ($\lesssim e^{\alpha_2/\alpha_3}$) is a constant representing the maximum effective number of tokens that can be learned from the first stage.
This modification is designed such that when $D_1 \gg D_{\rm max}$, $\hat{D_1}$ saturates to $D_{\rm max}$, preventing unbounded growth of the effective exponent.

\textbf{Large $D_2$ limit.}
Our scaling laws predict that base models trained with different number of tokens in the first stage ($D_1$) will converge at some point, and diverge again afterwards, with smaller $D_1$ models performing better at large $D_2$.
To see if the latter occurs (under reasonable compute constraints), we train a 15M model with different $D_1$ values, stack them with growth factor 2, and continue training with extremely large number of tokens in terms of token-to-parameter ratio (up to around a factor of 2000); see Figure \ref{fig:d1_dependence}, right panel.

However, we find that we effectively enter a noise floor at such limits, where stochastic noise can dominate the results.
To disentagle such effects, we decay the checkpoints multiple times with different random seeds to estimate the noise variance (details in Appendix \ref{app:extreme}). 
Despite these measures, the data remained too noisy to yield a definitive conclusion regarding the post-crossover regime.
% We find that too noisy to make any conclusion. 
We also do not observe crossover for a larger 0.5B-to-1B model; see Appendix \ref{app:extreme}. 
%  To summarize, we do not find clear empirical evidence that our scaling law holds after crossover.
%  Due to noise and exponentially more data required to explore the region after crossover, we leave more compute-heavy analysis to future work.
In summary, while our scaling laws predict a crossover, we do not yet find clear empirical evidence that they hold once this point is passed. Given the noise and the exponentially larger compute required to explore this territory, we leave this significantly more compute-intensive study as an open problem.
% We also notice a slight underestimation of the loss by the scaling law at very large $D_2$ values (in terms of token-to-parameter ratio); see Figure \ref{fig:extreme} in Appendix \ref{app:extreme}.
% This may be due to the limitation of the power-law ansatz, as also observed in \citet{hoffmann2022training}.
% Otherwise, our scaling laws fit the data well across a wide range of $D_1$, $D_2$ and $N$ values studied.

\section{Towards Understanding and Mitigating the Saturation Effects}
% We begin with studying mechanistic origins of the saturation phenomenon observed in model reuse.
% We show that the saturation phenomenon correlates with gradient norms, leaving other negative observational results in Appendix \ref{app:mechanistic}.

% \textbf{Saturation phenomenon correlates with gradient norms.}
\subsection{Mechanistic Causes}
% Here, we provide a mechanistic explanation of the saturation phenomenon observed in model reuse.
To understand the optimization dynamics of overtrained models, we analyze gradient norms during the second stage of pretraining.
% , leaving other negative observational results in Appendix \ref{app:mechanistic}.
%  We find that the saturation phenomenon correlates strongly with these norms.
We show in the left panel of Figure \ref{fig:valid} the gradient norm curves of overtrained and well-trained models during model reuse.
We find that overtrained models exhibit larger gradient norms generally, providing a form of mechanistic explanation for the saturation phenomenon.
This aligns with some previous findings \cite{klein2024plasticity}, and also aligns with our initial intuition that the over-optimized parameters are situated in a highly steep/sensitive part of the parameter space relative to the second-stage training, making further optimization difficult.
% We further study this phenomenon through the variance of training curves in Appendix ...
We have also investigated other possible observables but do not find strong correlation with overtraining; see Appendix \ref{app:mechanistic}.
\subsection{Practical Mitigation Strategy}

  Regularization techniques like gradient clipping and layer normalization, are effective at controlling gradient norms, which could potentially mitigate saturation effects.
 However, such techniques are already employed in our models.
  Instead of pursuing further optimization or architecture modifications, which may be impractical due to the complexity of large-scale training of language models, we adopt a \textbf{direct, data-driven approach}: leveraging the derived scaling laws for both second-stage and from-scratch pretraining to guide practitioners in minimizing the negative effects of overtraining.
   We specifically target the stacking-based model growth scenario, as it exhibits more serious saturation effects and training from scratch is a common practical alternative.

% \citet{lyle2023understanding} also showed that regularization, particularly layer normalization, is most effective at mitigating plasticity loss.
% However, layer normalization is already used in our models, along with other regularization techniques such as weight decay and gradient clipping.
% Instead of pursuing other optimization or architecture-modifying mitigation strategies, we choose a more direct approach, that is, leveraging the scaling law for bootstrapped and from-scratch pretrainings to provide guidance to practitioners for choosing training strategies that minimize the negative effects of overtraining.

% We target the scenario of stacking-based model growth, which has more serious saturation effects and from-scratch training is more common in practice.
Denote $L^{\rm scratch}_{2N}(D)$ by the loss of a model of size $2N$ trained from scratch for $D$ tokens, and $L^{\rm growth}_{N}(D_1,D_2)$ by the loss of a model of size $2N$ grown from a base model of size $N$ trained for $D_1$ tokens, and then trained for $D_2$ tokens.

In Figure \ref{fig:valid} (left panel), we show the plots of the two losses, $L^{\rm growth}_{N}(D_1,D_2)$, $L^{\rm scratch}_{2N}(D)$ with $D=D_2$, using our experimental data as well as the fitted scaling laws for 0.5B-to-1B grown model and from-scratch 1B model training.
We see that for small $D$ values, model growth outperforms from-scratch training for all $D_1$ values considered.
However, as $D$ increases, the losses from model growth saturate, and from-scratch training eventually outperforms it.
% Henceforth, whether one should choose model growth or from-scratch training depends on the $D_1$ value used in model growth, as well as the target $D_2$ or $D$ value.
% This simple strategy based on the scaling laws can help practitioners mitigate saturation effects in model reuse
Consequently, the optimal choice between model growth and from-scratch training depends directly on the initial investment ($D_{1}$) and the total target training tokens ($D_{2}$ or $D$).
 This analysis provides a simple, scaling-law-based strategy for practitioners to avoid the negative consequences of reusing overtrained base models.

\subsection{A Toy Model}
% While our results provide a mechanistic view via gradient norms, we seek a conceptual framework to explain why the second-stage scaling exponent decreases logarithmically with $D_1$.
Our previous results are mainly observational and we are unaware of existing theories \cite{bahri2024explaining,paquette20244+,maloney2022solvable} that can directly explain why the second-stage scaling exponent decreases logarithmically with $D_1$.
We henceforth take an alternative view by proposing a minimal toy setup in which the logarithmic saturation of the scaling exponent emerges, drawing on the work of learning curve theory \cite{hutter2021learning}, maximum entropy principle \citep{mandelbrot1953informational,jaynes1957information,jaynes1957information2}, and power-law assumptions. 
Detailed derivations are relegated to Appendix \ref{app:theory}.
% We extend the "learning curve theory" \cite{hutter2021learning}, which models the loss as the number of unseen examples with Zipf-distributed frequencies.

We model the feature space (of infinite dimension) as a $k$-ary tree with branching factor $k$.
Each datum consists of a single feature, and correct learning of a feature requires observing it once \cite{hutter2021learning}.
The rank $r$ of a feature at depth $d$ is approximately $r \approx k^d$.
We assume the cost of learning a feature at depth $d$ is proportional to $d \approx \frac{\log r}{\log k}$.
Then, following maximum entropy principle, the probability of finding a feature at depth $d$ decays exponentially ($\lambda > 0$ is a constant): $$P(d) \propto e^{-\lambda d}\propto r^{-\left( \frac{\lambda}{\log k} \right)},$$ establishing that the power-law exponent is $\alpha = \frac{\lambda}{\log k}$.

\citet{hutter2021learning} showed that if the data distribution follows power law with exponent $\alpha+1$, then the overall learning curve also follows a power law with exponent $\alpha$ for small $\alpha$'s.
This sets the exponent of learning with $D_2$.
To model the effect of $D_1$, we posit that first-stage pretraining uncovers more features, effectively expanding the effective branching factor $k$ by a power law: $k_{new} \approx k \cdot D_1^{\delta}$, where $\delta > 0$ is a constant.
Then, the new exponent becomes
$$\alpha_{new} \approx \frac{\lambda}{\log k + \delta \log D_1} \approx \alpha - \theta \log D_1,$$ with $\theta = \frac{\lambda\delta}{\left(\log k\right)^2}$.
% The structural learning model thus provides a theoretical parallel to our empirical law, i.e., logarithmic degradation of the scaling exponent.
We have thus shown (in a minimal setup) that saturation in the scaling exponent can emerge from a tree model with its branching factor modified by first-stage training.

% \textbf{Direct strategy to mitigate saturation effects.}

\section{Conclusion}
% In this paper, we have studied the scaling behavior of model reuse methods.
% In this paper, we have shown that while model reuse methods can accelerate learning, they exhibit scaling saturation as the base model is overtrained.
% Our work bears practical implications as even relatively tiny models are overtrained to trillions of tokens nowadays \citep{zhang2024tinyllama,dubey2024llama}, and model reuse is widely used in practice.
% We encourage the community to report the amount of pretraining data used for the base model, and even make intermediate checkpoints available, which can be more useful than the final model for model reuse.

We have provided a fairly broad scaling study of two-stage pretraining and there are other potential future directions worth diving deeper into, besides the ones we preliminarily studied in the main text (post-crossover region, mechanistic explanation, theory for saturated scaling): (1) Incorporating other factors into the scaling laws, e.g., replay ratio \citep{que2024d,wang2025learning} and the growth factor in model growth \citep{du2024stacking}. (2) Developing more advanced and scalable techniques to completely eliminate saturation effects, complementing the scaling-law-based guidelines presented in this paper.

\section*{Impact Statement}

This paper works toward the goal of advancing the field of machine learning and language modeling, with an emphasis on scaling.
There are many potential societal consequences of our work, none which we feel must be specifically highlighted here.


\bibliography{ref}
\bibliographystyle{icml2025}

%%%%%%%%%%%%%%%%%%%%%%%%%%%%%%%%%%%%%%%%%%%%%%%%%%%%%%%%%%%%%%%%%%%%%%%%%%%%%%%
%%%%%%%%%%%%%%%%%%%%%%%%%%%%%%%%%%%%%%%%%%%%%%%%%%%%%%%%%%%%%%%%%%%%%%%%%%%%%%%
% APPENDIX
%%%%%%%%%%%%%%%%%%%%%%%%%%%%%%%%%%%%%%%%%%%%%%%%%%%%%%%%%%%%%%%%%%%%%%%%%%%%%%%
%%%%%%%%%%%%%%%%%%%%%%%%%%%%%%%%%%%%%%%%%%%%%%%%%%%%%%%%%%%%%%%%%%%%%%%%%%%%%%%
\newpage
\appendix
\onecolumn
\section{Notations}
\label{app:notation}
 We summarize main notations used in the paper.
\begin{itemize}
    \setlength{\itemsep}{0pt}      
    \setlength{\parskip}{0pt}      
    \setlength{\itemindent}{0pt}   
	\setlength{\labelsep}{4pt}     
    \item $L$: Cross-entropy loss or the (natural) logarithmic loss
    \item $D$: Dataset size in token
    \item $N$: Non-embedding model size or number of non-embedding parameters
    \item $A,B,F$: Scaling factors of the power law, independent of the variable under consideration
    \item $E$: Irreducible loss of the power law, independent of the variable under consideration
    \item $\alpha,\beta,\gamma$: Scaling exponents of the power law, independent of the variable under consideration
    \item $n_{\rm layer}$: number of layers of a model
    \item $d_{\rm model}$: hidden dimension size of a model
    \item $d_{\rm MLP}$: intermediate hidden dimension size of a model
\end{itemize}

\section[short]{More on Architecture and Experimental Design}
\label{app:exp}
\subsection{Megatron-LM Configuration}
\textbf{Infrastructure.}
Our experiments are performed on multiple nodes, each consisting of 8 NVIDIA H100 80 GB GPUs, interconnected via InfiniBand HDR.
The software we use for training is the Megatron-LM library \citet{shoeybi2019megatron}.

We use and modify the Megatron-LM (core v0.8.0) library for our experiments\footnote{\url{https://github.com/NVIDIA/Megatron-LM}}.
Models are trained with data type bfloat16.
% Except for the largest MoE we train (8x1B), which has tensor parallelism configured to be 2, all models are trained with data and sequence parallelisms only \citet{korthikanti2023reducing}.
Other optimization libraries used include FlashAttention \citet{dao2022flashattention} and TransformerEngine\footnote{\url{https://github.com/NVIDIA/TransformerEngine}}.
See the example scripts provided in supplementary material. 

\subsection{Model Configuration}
Let us elaborate more on our model configuration.
The intermediate hidden dimension size, $d_{\rm MLP}$, is set to be four times the hidden dimension size, i.e., $4d_{\rm model}$.
Bias is not used in the linear layers.
We do not consider efficiency-motivated implementations like grouped query attention as well.
Attention head number is chosen to increase with model size following standard practices.
Other designs of the architecture follow Llama2's closely \citep{touvron2023llama}.

We vary model sizes keeping the ratio $n_{\rm layer}/d_{\rm model}$ to lie in the range 32 to 64, as in \citet{kaplan2020scaling}.
Smaller models for used for ablation studies.
See Table \ref{tab:config} for the exact numbers for the model configuration.

\begin{table}[htb]
   \caption{Models used in our study and their parametric details. 
    Note that $d_{\rm MLP}$, is set to be $4d_{\rm model}$.}
    \label{tab:config}
    \centering
    \begin{tabular}{l|cccr} 
        \toprule
        Model  & $n_{\rm layer}$ & $d_{\rm model}$ & $n_{\rm head}$ & $N$ \\
        \midrule
        \textbf{15M} & 9  & 320  & 4 &14,751,680
        \\
        \textbf{44M} & 12  & 480  & 8 &44,248,800\\
        \textbf{0.1B}  & 15 & 640  & 8 &98,323,840 \\
        \textbf{0.2B}  & 21 & 832 & 8 &232,623,040 \\
       \textbf{0.5B}  & 26 & 1,120 & 16 &521,889,760 \\
        \textbf{1B}& 30& 1,504& 16&1,085,859,424\\
        \bottomrule
    \end{tabular}
    \end{table}
\subsection{Training Configuration}
% The common setup of training is shown in Table \ref*{tab:common}, and the model-dependent setup (warmup iteration, standard deviation of the normal distribution for initializing model parameters, maximum iteration run, batch size, tuned LR) is shown in Table \ref{tab:train_config}.
% As described in the main text, we use the WSD schedule for training.
% The number of warmup steps of the WSD LR schedule is set to be roughly the same as the total model size \citet{porian2024resolving}.
% Linear decay to 10\% of the maximum LR value is used in the last stage of the schedule, with the length set to be around 10\% of the training length, following \citet{hagele2024scaling}.

% It is noted that we do not use techniques geared for treating training instabilities (which we did not encounter in our study), such as Z-loss or QK normalization. 

% Logarithmically-spaced intermediate checkpoints are saved and used to emulate different numbers of training token budget. 
% We also increase both the training length and batch size with model size following common practices without performing precise tuning. 

As discussed in the main text, we adopt the WSD learning rate schedule for all experiments.
The number of warmup steps is set approximately equal to the model size, as suggested by \citet{porian2024resolving}.
In the final phase of training, the learning rate decays linearly to 10\% of its peak value, with the decay phase spanning roughly 10\% of the total training steps, following the setup in \citet{hagele2024scaling}.
To emulate varying token budgets, we save intermediate checkpoints at logarithmically spaced intervals.

The general training configuration is summarized in Table~\ref{tab:common}, while model-specific hyperparameters, such as warmup iterations, initialization standard deviation ($\sqrt{2/5d_{\rm model}}$ \citep{le-scao-etal-2022-language}), maximum training steps, batch size, and tuned learning rate, are provided in Table~\ref{tab:train_config}.
Batch size is scaled with model size according to standard practice, without tuning for optimality.
The number of tokens is increased with model size as well, such that the parameter-to-token ratio remains roughly constant across different model sizes.
We note that no specialized techniques for mitigating training instabilities, such as Z-loss or QK normalization, are employed, as such instabilities do not arise in our experiments.

\begin{table*}
   \caption{Training configuration used throughout the paper.}
    \label{tab:common}
    \centering
    \begin{tabular}{l|l}
    \hline
    Configuration             & Details \\ \hline
    Context length & 1,024 \\
    Embedding tying & False \\
    Optimizer & AdamW \citet{loshchilov2017fixing} \\    
    Adam $\beta_1$                   & 0.9 \\
    Adam $\beta_2$                & 0.95 \\
    Adam $\epsilon$  & 1e-8 \\
    Weight decay                 & 0.1 \\
    Gradient clipping   & 1.0 \\
    \end{tabular}
    \end{table*}

\begin{table*}[htb]
    \centering
     \caption{\textbf{Model-dependent training configuration}. 
    "init. size" refers to the standard deviation of the normal distribution used for initializing the weights.
    "Std iter." refers to the number of iteration run on the model to reach about 70 times the model size in tokens, the standard training length used in the scaling law experiments. For several models, we also run longer training for exploration.}
    \label{tab:train_config}
    \begin{tabular}{l|ccccc} 
        \toprule
        Model & warmup iter. & init. size & Std iter. & batch size &LR\\
        \midrule
        \textbf{15M} & 200  & 0.035  & 17,600& 128&8e-3  
        \\
        \textbf{44M}  & 200  & 0.029  & 17,600& 256&4e-3  \\
        \textbf{0.1B}   & 200 & 0.025  & 17,600 &512&4e-3   \\
        \textbf{0.2B}  & 400 & 0.022 & 35,200 &512&2e-3   \\
        \textbf{0.5B}  & 800 & 0.019 & 70,400 &512&4e-4   \\
        \textbf{1B}& 800& 0.016& 70,400&1024&4e-4 \\
        \bottomrule
    \end{tabular}
   
    \end{table*}

\subsection{On the choice of Data Domains Studied}
{
Here, we argue that our experimental setup for CPT incorporates a significant domain shift, and it is highly relevant to frontier model research.

\begin{itemize}[leftmargin=*]
    \item \textbf{Domain gap in our setup.}
The base models are purposefully pretrained exclusively on the Common Crawl portion of the Slimpajama-DC dataset.
 This explicitly excludes data heavily skewed toward specialized domains such as mathematics (like arXiv) or code (like dedicated GitHub repositories), thereby maximizing the domain gap for the subsequent CPT stage.
    \item \textbf{Small overlap of domains.}
While some minimal token overlap between the generic web corpus and the specialized domains is inevitable, it is quantitatively small.
 We estimate that math/code data accounts for approximately 0.3 percent of the total Slimpajama-DC corpus. \footnote{Estimated based on filtering performed for the OpenCoder \citep{huang2025opencoder} dataset, which extracted math/code data from a similar CommonCrawl-derived corpus, FineWeb \citep{penedo2024fineweb}.}
Consequently, the domain shift in our CPT experiments, which move to dedicated code and math corpora, is substantial.
    \item \textbf{Relevance to frontier LLM training.}
The strategy of pretraining mainly on generic web data and subsequently applying CPT on specialized domains like code (Stack/StarCoder) and mathematics (OpenWebMath) is a standard and highly relevant practice for state-of-the-art LLMs.
    \item \textbf{Model growth.}
For model growth experiments, the primary goal is to isolate and quantify the effect of increasing model capacity, not domain shift.
 Therefore, the second stage is intentionally conducted on the same generic web dataset as the base model, allowing us to accurately quantify the scaling behavior of the growth technique itself.
\end{itemize}

 To summarize, our current findings are robust within the most practically relevant and challenging domain-transfer scenarios currently employed for large-scale LLM training.}
\subsection{GPU Hours and Costs}
\label{app:gpu}
Instead of reporting the actual runtimes on our cluster, which varied in our experiments due to many factors affecting the cluster (number of available nodes, congestion, etc.), we give a theoretical estimate of total GPU hours used for obtaining the joint scaling law, which involves running the largest tested model with most training tokens in this paper.

The estimate is as follows.
We calculate the FLOPs for training the largest first and second-stage models with maximum iterations using the $6ND$ approximation, ignoring the additional FLOPs required to continued pretrain models with shorter iterations (as we can reuse the intermediate checkpoints). \footnote{We have more than 450 runs; estimated from 25 data points for each of the 8 methods considered in Table \ref{tab:func_flipped}, and 125 data points for the two joint scaling law experiments (CPT and model growth).}
We further assume that the per-second TFLOPs of the GPU is 400.
We obtain around $10^{9}$ TFLOPs. 
Taking into account of additional experimentation and ablation runs, we estimate that around 3000 GPU hours were used for the entire project.
Assuming a cost of 2 USD per GPU hour, the total cost is around 10,000 USD.

{
\textbf{Scaling law methodology and cost justification.} We note that our study contains over 450 total runs, as accurately mapping the relationship between multiple scaling variables (model size $N$, first-stage tokens $D_{1}$, and second-stage tokens $D_{2}$) necessitates a multitude of controlled experiments on models of various sizes.
 This methodology contrasts with single, large-scale training runs (e.g., training one 7B model from scratch), which would provide fewer data points for fitting complex scaling laws. 
 This rigorous approach, balanced with the need for a reasonable overall budget, dictates our focus on models up to 1B parameters for extensive experimentation.
}
    \section{Training Details}
    \label{app:train}
    \subsection{Ablation of Learning Rate Schedules}
\label{app:schedule}
\begin{figure*}[h]
    \centering
        \includegraphics[width=0.4\linewidth]{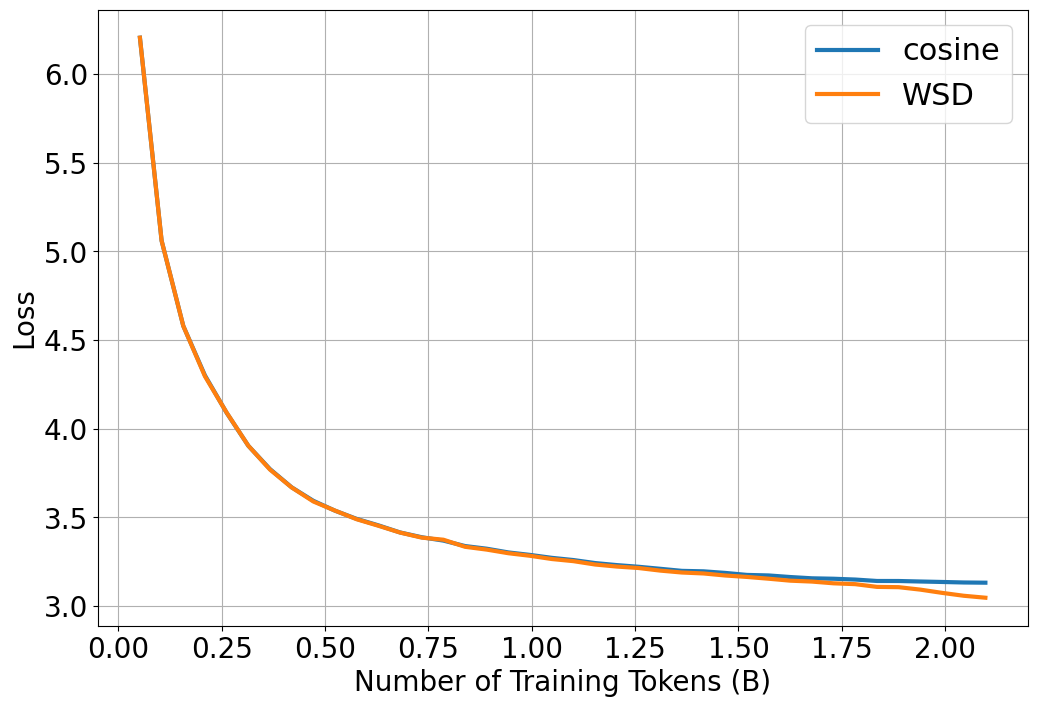}
        \includegraphics[width=0.4\linewidth]{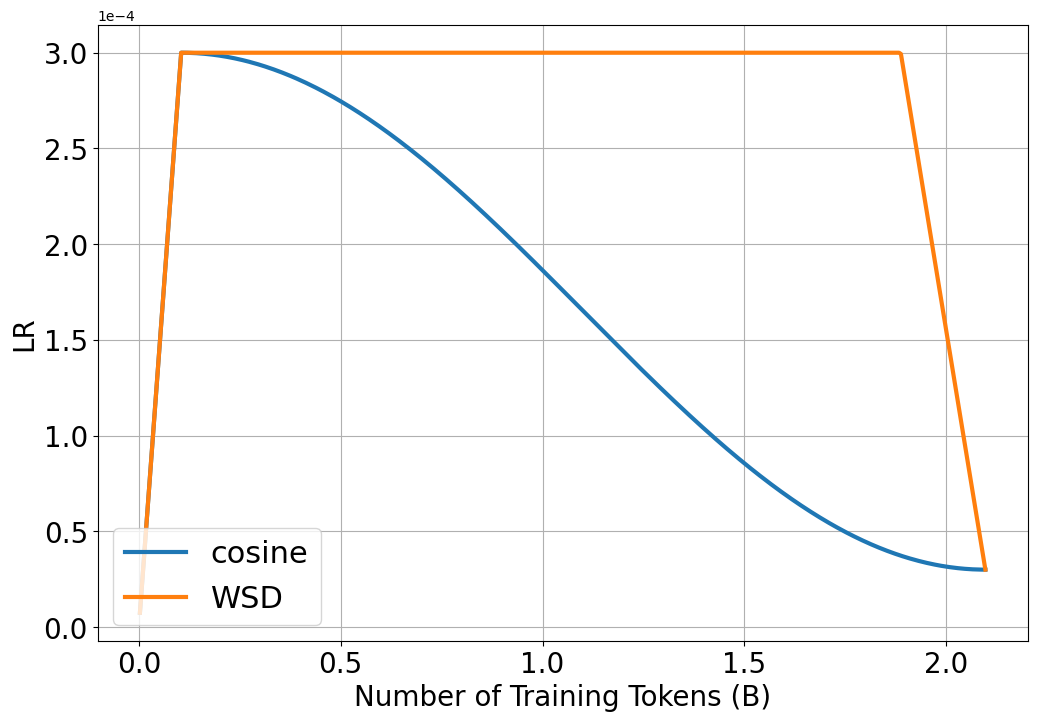}
    \caption{
        \textbf{WSD vs cosine LR schedule.} 
        We show that the WSD LR schedule achieves similar (even slightly better) final loss compared to the more commonly used cosine LR schedule.
        }   \label{fig:ablate_wsd}
\end{figure*}

Here, we compare the performances of using WSD and the commonly used learning rate (LR) cosine schedules (decaying to 10\% of the peak LR value) \citep{touvron2023llama}.
The model size we use is 0.1B, with the following training configuration: batch size 512, 4,000 training iterations, and 200 warmup iterations.
We can see from Figure \ref{fig:ablate_wsd} that both schedules yield similar (with WSD achieving slightly better final loss) performances.
This justifies our choice of using the WSD schedule throughout the paper.
    % \begin{table*}
    %     \centering
    %      \caption{Training configuration for ablation studies.}
    %     \label{tab:schedule}
    %     \begin{tabular}{l|l}
    %     \hline
    %     Configuration             & Details \\ \hline
    %     Batch size & 512 \\
    %     train iter.                    & 4,000 \\
    %     Warmup iter.               & 200 \\
    %     \hline
    %     \end{tabular}
    %     \end{table*}

\subsection{Ablation of Data Repetition}
\label{app:repetition}
\begin{figure*}[h]
    \centering
        \includegraphics[width=0.4\linewidth]{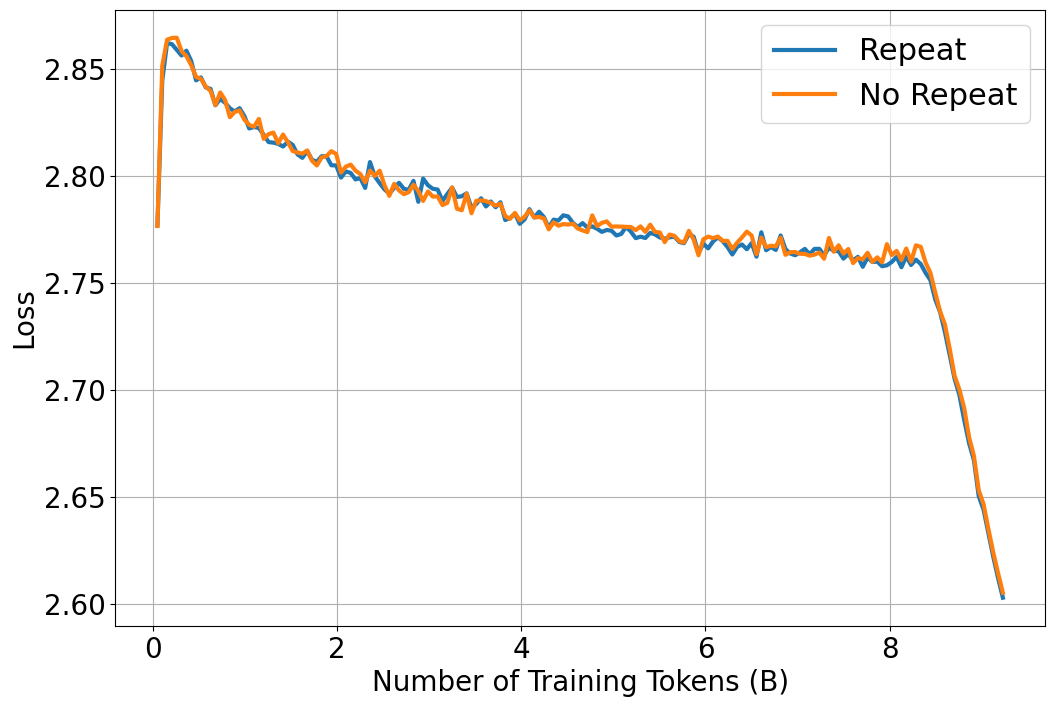}
    \caption{
        \textbf{Ablation of data repetition during model growth.} 
        We compare the performance of using the same data as the base model and using different data for the second stage of training.
        The difference is small, indicating that data repetition does not affect the scaling behavior significantly.
        }   \label{fig:ablate_repetition}
\end{figure*}

During model growth, we have used the same data as the base model for the second stage of training.
This is because the base model is often trained with a large amount of data, and it is not always possible to obtain a large amount of additional data.
However, this means that the second stage of training involves repeating the same data.
To study the effect of data repetition, we have made an ablation study, where we compare the performance of using the same data as the base model and using different portion of the Slimpajama data. 
As shown in Figure \ref{fig:ablate_repetition}, we find that the difference is small.
\subsection{Evaluation with Standard Benchmarks}
\label{app:eval}
We compare the performance of our trained 1B base model against existing models with similar sizes, Pythia \citep{biderman2023pythia} and TinyLlama \citep{zhang2024tinyllama}, based on standard natural language processing benchmarks, ARC \citep{clark2018think}, lambada \citep{paperno2016lambada}, logiqa \citep{liu2020logiqa}, piqa \citep{bisk2020piqa}, sciq \citep{welbl2017crowdsourcing}, and winogrande \citep{sakaguchi2021winogrande}.

Table \ref{tab:eval} shows the results.
We see that the model perform similarly to the open models, indicating that our models have been trained correctly.

\begin{table}[h!]
    \centering
    \caption{\textbf{Benchmarks' performance comparison across models.}
    Reported scores are accuracies (normalized by byte length whenever applicable).
    The first two columns are scores of existing models.
    The last column is evaluation results of the largest base model trained with the most number of tokens in this work.
    Our models are evaluated with the LM Evaluation Harness v0.4.0 library \citep{eval-harness}.}
    \label{tab:eval}
    \begin{tabular}{lcc|c}
    \toprule
    \textbf{Models} & \textbf{Pythia-1B} & \textbf{TinyLlama-1.1B} & \textbf{Our 1B model}  \\
    \midrule
    \textbf{Datasets} & Pile & Slimpajama \& Starcoder & Slimpajama  \\
    \textbf{Tokens}& 100B & 103B & 74B  \\
    \midrule
    ARC-c        & 25.59 & 24.32 & 27.65  \\
    ARC-e        & 47.26 & 44.91 & 52.10 \\
    lambada      & {53.52} & -     & 45.08 \\
    logiqa       & {29.49} & -     & 26.11 \\
    piqa         & {69.31} & 67.30 & 65.89 \\
    sciq         & 77.3  & -     & 78.10  \\
    winogrande   & 51.22 & 53.28 & 54.93 \\
    \midrule
    \textbf{Avg.} &50.53 & -     &49.98  \\
    \bottomrule
    \end{tabular}
    \end{table}

\subsection{Continual Pretraining Details}
\label{app:cpt}
Using a lower LR for CPT is also a common practice \citep{ibrahim2024simple}.
As an ablation, we have experimented with using a constant LR schedule setting the value to be the final LR for the first-stage base model training, on a 0.1B model.  
We find that it yields worse performance (see Figure \ref{fig:cpt_lr}).
Hence, we use the same LR (and LR schedule) as the base model for CPT.
\subsection{Model Growth Details}
\label{app:mg}
For completeness, we provide more details on the model growth methods used in this work.

\textbf{Width expansion details.}
Let us elaborate more on the width expansion method used in this work.
We first give a more precise defintion of function preservation.
Let $F$ be a function and $G$ as the growth operator.
The function preservation condition is defined as:
\begin{equation*}
    F(x)=G(F)(x), \forall x \in \mathcal{X
}
\end{equation*}
where $\mathcal{X}$ is the input space.

When performing width expansion, the neuron values of each layer are expanded by duplicating the weights of existing neurons, and dividing the output weights by the growth factor.

\textbf{Stacking details.}
We first note that stacking does not preserve function but is empirically found to work well in practice \citep{du2024stacking}.
We use the recommended stacking procedure in \citet{du2024stacking}, which is, letting $M$ be the non-embedding part of the base model, the stacked model with growth factor $k$ is given by:
\begin{equation*}
    M'  = M \circ M \circ ... \circ M \quad (k \text{ times})
\end{equation*}
where $\circ$ denotes function composition.
The embedding and final layer are then simply copied from the base model.

\begin{figure*}[h]
    \centering
        \includegraphics[width=0.4\linewidth]{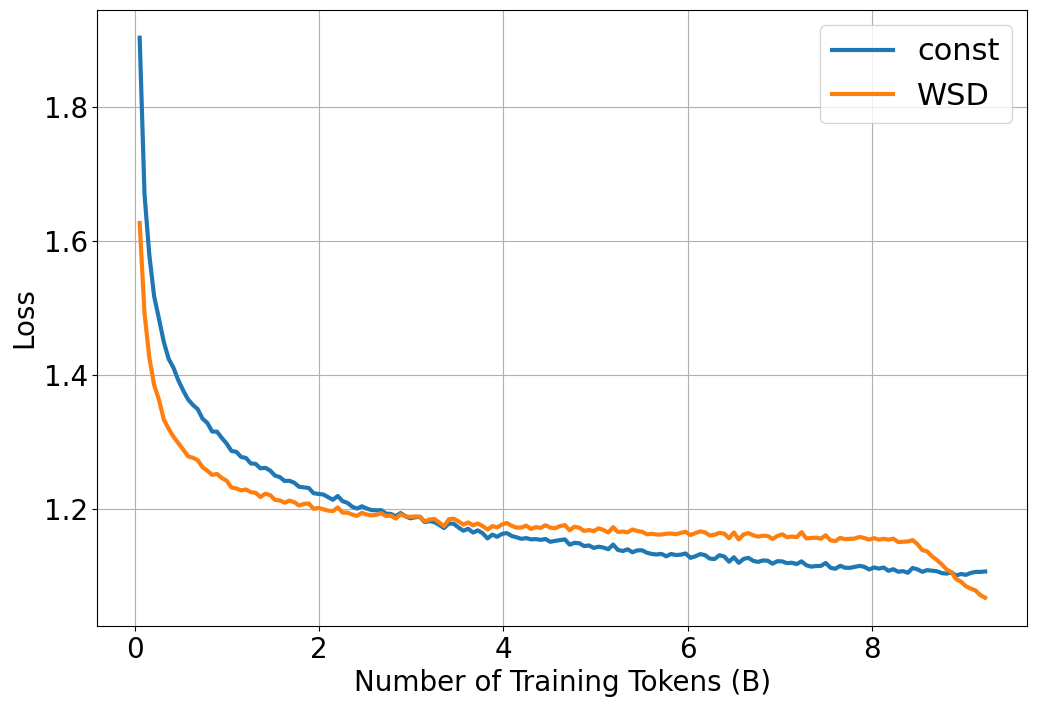}
        \includegraphics[width=0.4\linewidth]{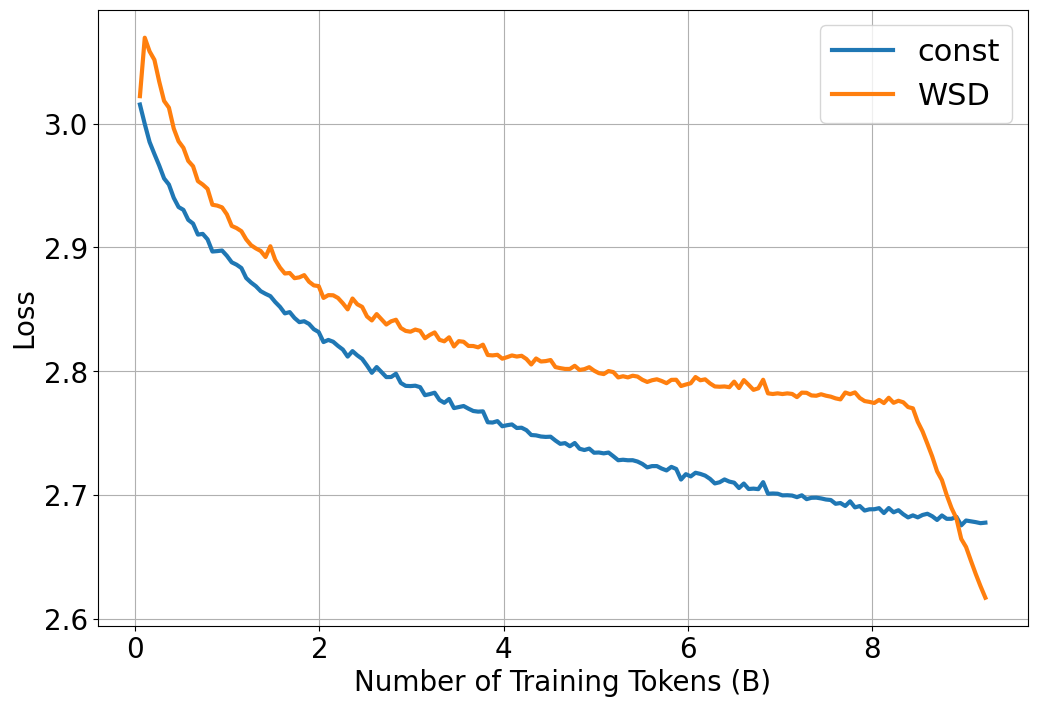}
    \caption{
        \textbf{model reuse with lower LR.} 
        We show that using the same LR as the base model achieves better final performance than using a lower LR for a 0.1B model continual pretrained on code data (left) and stacking (right).
        }   \label{fig:cpt_lr}
\end{figure*}

\textbf{Low LR.}
We also experiment with using a lower constant LR for stacking as above, and find that it also yields worse performance (see Figure \ref{fig:cpt_lr}).
Hence, we use the same LR (and LR schedule) as the base model for stacking as well.

\section{Scaling Laws Details}
\label{app:scaling}
\subsection{Fitted Exponents and Other Results}
We show the fitted exponents of the multiplicative scaling laws studied in Table \ref{tab:func_flipped} in Table \ref{tab:func_2_exponents}.

\begin{figure*}[h]
    \centering
        \includegraphics[width=0.45\linewidth]{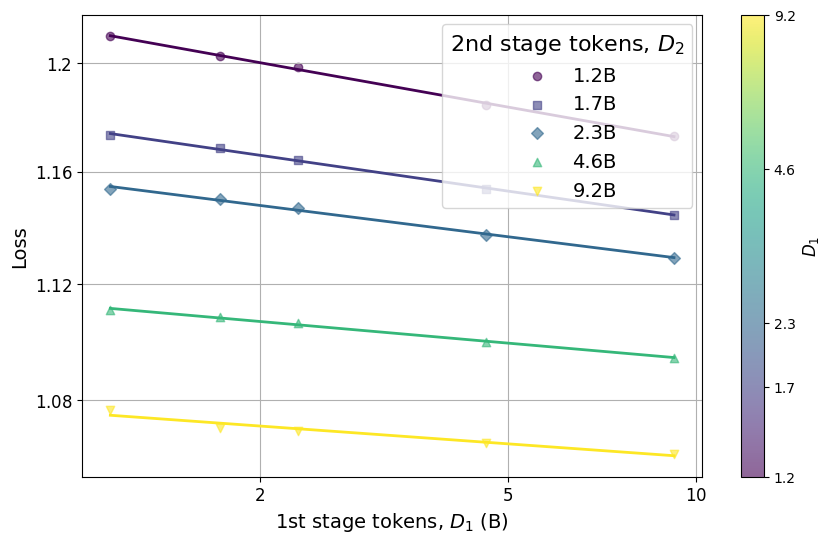}
        \includegraphics[width=0.45\linewidth]{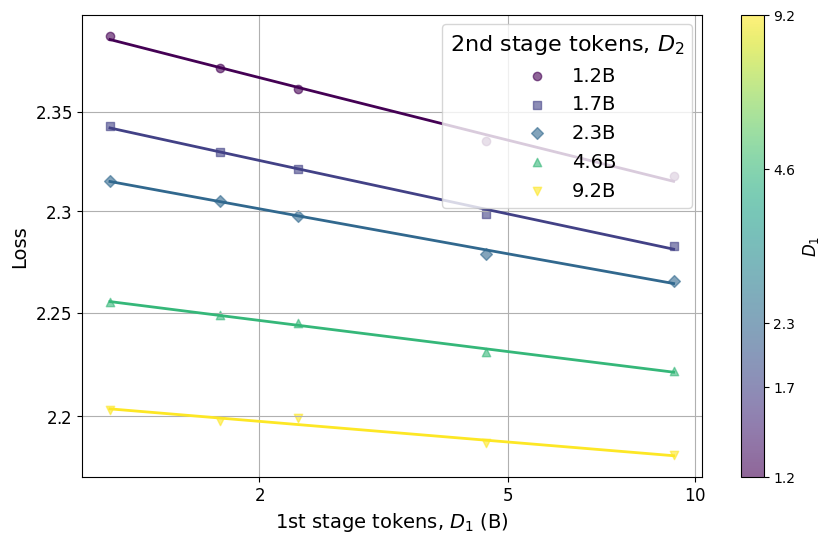}
        \includegraphics[width=0.42\linewidth]{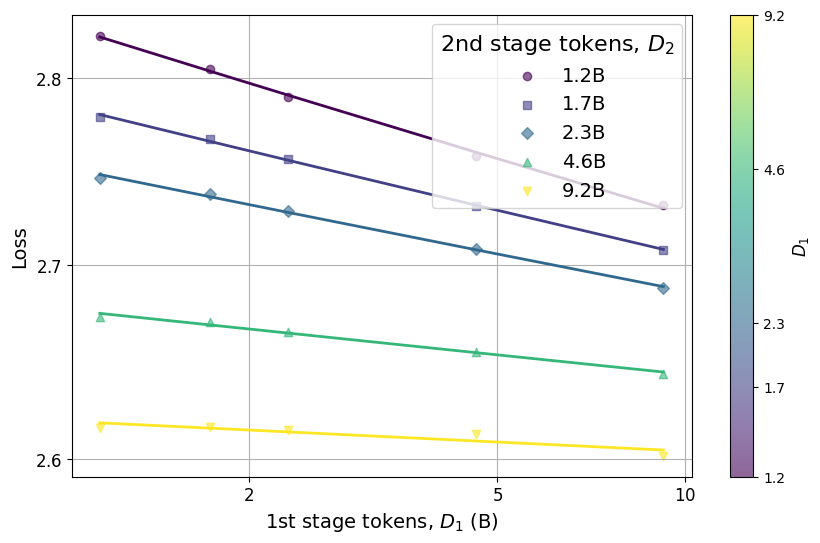}
        \includegraphics[width=0.45\linewidth]{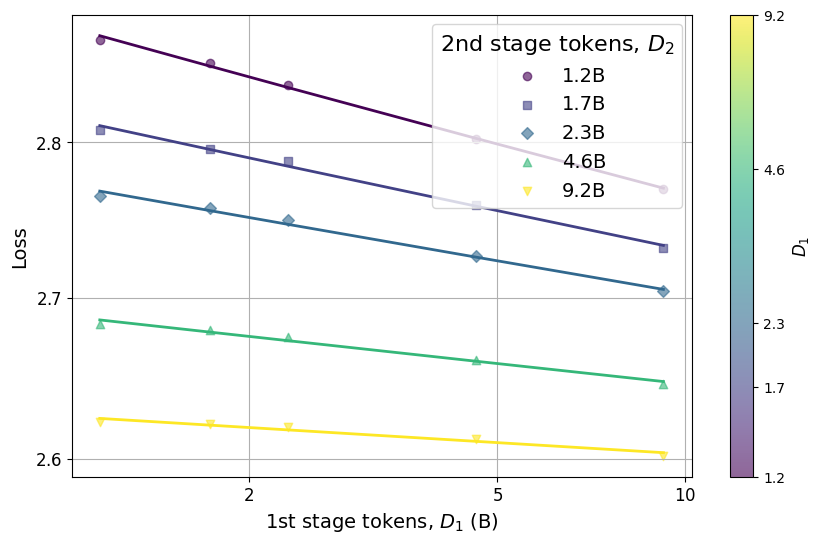}
    \caption{
        \textbf{ $D_1$ has power-law scaling.} 
        We show scaling behavior of fist-stage training tokens ($D_1$) for different values of second-stage tokens ($D_2$), indicating that $D_1$ also has power-law scaling.
        \textbf{Top left:} Continual pretraining on code. \textbf{Top right:} Continual pretraining on mathematics data. \textbf{Bottom left:} Model growth by stacking (growth factor 2). \textbf{Bottom right:} Model growth by expansion (growth factor 2).}   \label{fig:d1_dependence2}
\end{figure*}

We further show that $D_1$ has power-law scaling in Figure \ref{fig:d1_dependence2}, for CPT on mathematics data and model growth by expansion, which justifies the multiplicative form of the scaling laws.

\begin{table}[htb]
    \caption{
    \textbf{Fitted exponents for the multiplicative scaling law.} 
    Corresponding \(\alpha_1, \alpha_2, \alpha_3\) values for the multiplicative interaction fits shown in Table \ref{tab:func_2_exponents}.
    }
    \label{tab:func_2_exponents}
    \centering
    \begin{tabular}{l|ccc}
        \toprule
        Variant & $\alpha_1$ & $\alpha_2$ & $\alpha_3$ \\
        \midrule
        \textbf{CPT (code)} & 0.106 & 0.146 & 0.004 \\
        \textbf{CPT (math)} & 0.981 & 0.388 & 0.017 \\
        \textbf{Expand} & 0.549 & 0.852 & 0.024 \\
        \textbf{Stack} & 0.515 & 0.350 & 0.017 \\
        \textbf{CPT (replay)} & 0.424 & 0.626 & 0.018 \\
        \textbf{CPT (stable)} & 0.920 & 0.156 & 0.004 \\
        \textbf{Stack (x4)} & 0.644 & 0.829 & 0.028 \\
        \textbf{Stack (stable)} & 0.891 & 0.507 & 0.009 \\
        \bottomrule
    \end{tabular}
\end{table}

\subsection{Another variant of CPT scaling law.}
\label{app:cont}
When the dataset used for CPT is the same as the base model, we expect the following continuous scaling formula shown below holds:
\begin{equation}
    L(D_1,D_2)= A (D_1+D_2)^{-\alpha} + E 
    \label{eq:continuous} 
\end{equation}
Hence, one may expect this holds for CPT on a different dataset as well, as assumed in \citet{que2024d,wang2025learning}.
Fitting the above formula for the code (math) CPT scenario, we find that the RMS error is 0.0213 (0.0235), higher than the best ones in Table \ref{tab:func_flipped}.
Hence, our scaling law models CPT better, in contrast to previous assumptions, which may have overlooked the discontinuous (overtraining) effects of CPT.

One may further wonder if the model growth methods follow the above scaling law, since both stages use the same dataset.
However, we find that the RMS error for (growth factor 2) expansion and stacking methods are 0.032 and 0.025 respectively, again larger than the ones in Table \ref{tab:func_flipped}.
Even though the same dataset is used in training, the model capacity is changed in model growth, explaining why the continuous scaling law does not hold.
In Figure \ref{fig:valid_stack}, we show that the continuous scaling law does not fit the data well visually.

% \subsection{Comparing Functional Forms Visually}
% \label{app:other_funcs}

% We provide visual comparisons of different functional forms considered in Table \ref{tab:func_flipped}.
% We compare loss-versus-token plots of our proposed scaling law against other functional forms in Figure \ref{fig:other_funcs}.
% We see that the predicted losses of the alternative functional forms increasingly diverge from our proposed scaling law at large $D_2$. 
% Furthermore, the predicted losses of other functional forms with different $D_1$ are nearly parallel across the range of $D_2$, whereas ours converges towards each other to a crossover point, where 
% does not.
% This lack of parallelism accurately captures the saturation effect: as the base model is overtrained (higher $D_1$), the efficiency gain from additional $D_2$ tokens diminishes, leading to the lines converging at large $D_2$. 
% \begin{figure*}[h]
%     \centering
%         \includegraphics[width=0.6\linewidth]{interpolate_other_funcs.png}
%     \caption{
%         \textbf{Stacking with extremely large token-to-parameter ratio.} 
%          Conclusion similar to those made in Figure \ref{fig:sunk_cost} can be made for CPT on code data, where the scaling efficiency decreases with first-stage training tokens.}   \label{fig:other_funcs}
% \end{figure*}

\subsection{Joint Scaling Law}
\label{app:joint}
\begin{figure*}[h]
    \centering      
    \includegraphics[width=0.45\textwidth]{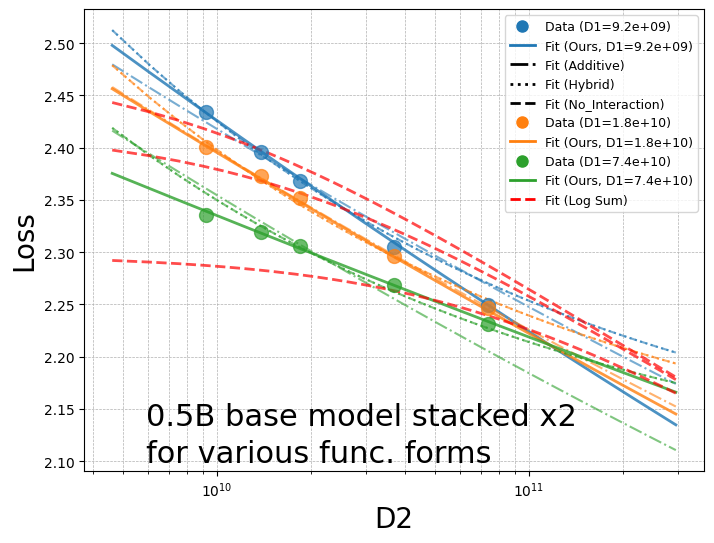}
    \includegraphics[width=0.45\textwidth]{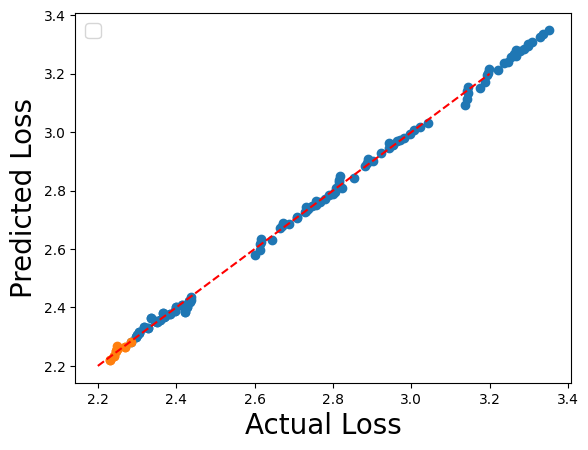}
    \caption{
    \textbf{Left: Fits with different functional forms.} We show in red the fits of the continuous scaling law does not capture the saturation effect well compared to our proposed multiplicative scaling law.    
    \textbf{Right: Fits of the joint upcycling scaling law of stacking.}
    10\% of the collected data points with lowest losses are used for validation (orange
    points).}
    \label{fig:valid_stack}
\end{figure*}

For model growth, we also fit the joint scaling law in Equation \ref{eq:joint}.
Note that there is ambiguity in defining the model size $N$ in the context of model growth.
We consider $N$ to be the size of the model before growth, and as we keep the growth factor fixed to be 2 in our experiments, the new model size after growth is $N'=2N$.
Therefore, $N$ and $N'$ differ by a constant factor of 2, which can be absorbed into the coefficient $B$ in Equation \ref{eq:joint}.
Unless stated otherwise, we use $N$ in the fitting of the joint scaling law for model growth.
In Figure \ref{fig:valid_stack}, we show the fit of the joint scaling law for stacking.

We provide the fitted coefficients of the joint scaling law in Table \ref{tab:param}.
In addition to model reuse, we also show the fitted coefficients for base models trained from scratch on the same dataset as the second stage (code data for CPT, and the same data as the base model for model growth).
For model growth, we consider a growth factor of 2, and fit the parameters with model size $N$ before stacking for comparison conveniences. 
We further note that the fitted coefficients are produced by fitting the joint scaling law to \emph{all} data points collected, including those used for validation in Figure \ref{fig:valid_stack}.

\begin{table}[htb]
    \centering
        \caption{\textbf{Fitted coefficients for joint scaling laws.}
    Note that for stacked models, we fit the coefficients with model size $N$ before stacking for comparison conveniences.}
    % The irreducible loss applicable to first and second stages pretraining for CPT and model growth respectively, $E$'s, are fitted to be the consistent for each of the scenario.}
    \begin{tabular}{l|ccccccc} 
        \toprule
        {} & $\alpha/\alpha_1$ & $\alpha_2$ &   $\alpha_3$& $\beta$ & $A$ & $B$ & $E$ \\
        \midrule
        \textbf{Base (Slimpajama)} & 0.092&-&-&0.105& 10.383 &  10.085 & 0.041\\
      
        \midrule
        \textbf{From-scratch (code)} & 0.113&-&-&0.234& 8.143 &  27.286&0.105 \\
        \midrule
        \textbf{CPT (code)} & 0.048&0.126&0.001& 0.238 & 15.062 &  27.234& 0.105 \\
        \midrule
        % \textbf{Stack} &  0.043&0.075&0.001& 0.178 & 12.549 &  24.001& 0.041 \\
        \textbf{Stack} & 0.087&0.119&0.003& 0.173 & 33.394 &  22.471& 0.041 \\
        \bottomrule
    \end{tabular}
    % The irreducible loss applicable to all laws, $E$, is fitted to be 0.165.}
    \label{tab:param}
\end{table}
\subsection{Why Fitting Scaling Laws Separately?}
We justify our decision to fit two separate scaling laws—one for models trained from scratch and another for model reuse—instead of employing a unified formulation that spans all stages.

First, the scaling behavior of models trained from scratch is expected to differ from that of models undergoing second-stage pretraining. 
Specifically, at the limit $D_1 \to 0$, grown models are initialized from a pretrained base model, whereas scratch-trained models start from random weights. 
This difference in initialization leads to distinct learning dynamics and, consequently, different scaling law parameters.

Similarly, the scaling behavior of the base model differs from that of the second-stage training in the limit $D_2 \to 0$. As shown in Figure~\ref{fig:cpt_lr}, second-stage training often begins with a \emph{rewarming} phase, during which the loss initially increases before decreasing. 
This early instability deviates from the expected scaling of dense models, although the overall trend remains correlated—supporting the validity of Condition~1, due to function preservation and empirical observations that rewarming does not entirely disrupt loss behavior (as quoted from \citet{clauset2009power}: ``In practice, few empirical phenomena obey power laws for all values of $x$'').

Finally, our preliminary experiments indicate that a unified scaling law does not provide a satisfactory fit across both stages. 
We therefore opt to model them separately.
 We also note that there may exist alternative functional forms that could better capture the full range of behavior, but they may violate some of the well-established conditions for scaling laws (power law), and may overcomplicate the analysis or overfit the data; hence, we leave their exploration to future work.

\section{More Observations and Implications of the Scaling Laws}
\label{app:more_observe}

\subsection{Implication based on Scaling Laws' Extrapolation}
\label{app:extrapolate}
We consider an implication based on the extrapolation of the scaling laws, motivated by studies on MoE upcycling \citep{komatsuzaki2022sparse,pmlr-v267-liew25a}.

A key practical consideration is how much of the \emph{initial investment} in pretraining, the so-called \emph{sunk cost} ($D_1$), can be effectively leveraged in the second stage of training. This question has been explored in the context of MoE upcycling strategies. 
For example, it was found that upcycling yields benefits up to 120\% of the sunk cost. That is, to match the performance of an upcycled MoE model that underwent an additional 0.4 trillion tokens of training after an initial 2T tokens, training a comparable MoE from scratch would require 2.4T tokens—representing an effective saving of 2T tokens.

% \liew{move this to appendix because referee might ask why no experimental validation; here we use a similar figure in previous paper}
% To further examine this trade-off, we fit Chinchilla-style scaling laws to models trained \emph{from scratch} on the same datasets used in our second-stage experiments—namely, code or math data for CPT, and the original dataset for model growth. In Figure~\ref{fig:sunk_cost}, we plot the validation loss as a function of training tokens for both approaches, across a range of sunk costs, using a fixed model size of 7B. For CPT on code data, we find that models trained from scratch begin to outperform bootstrapped models (pretrained on 10T tokens) once more than 300B tokens of domain-specific data are used.
We investigate this in model growth by stacking (growth factor 2).
 We define $D^*$ as the number of tokens required for training from scratch to match the performance of a grown model with the same sunk cost, following \citet{pmlr-v267-liew25a}:
\begin{equation}
L^{\rm scratch}_{2N}(D^*)=L^{\rm grown}_{N}(D_1=D^*,D_2=D^*) \label{eq:sol}
\end{equation}
where $L^{\rm scratch}_{2N}(D)$ is the loss of a model of size $2N$ trained from scratch for $D$ tokens, and $L^{\rm grown}_{N}(D_1,D_2)$ is the loss of a model of size $2N$ grown from a base model of size $N$ trained for $D_1$ tokens, and then trained for $D_2$ tokens.
To obtain $L^{\rm scratch}_{2N}(D)$, we fit the losses of models trained from scratch with the Chinchilla-style scaling law (see Table \ref{tab:param} in Appendix \ref{app:joint} for the fitted coefficients).

\begin{figure*}[h]
    \centering
        \includegraphics[width=0.45\linewidth]{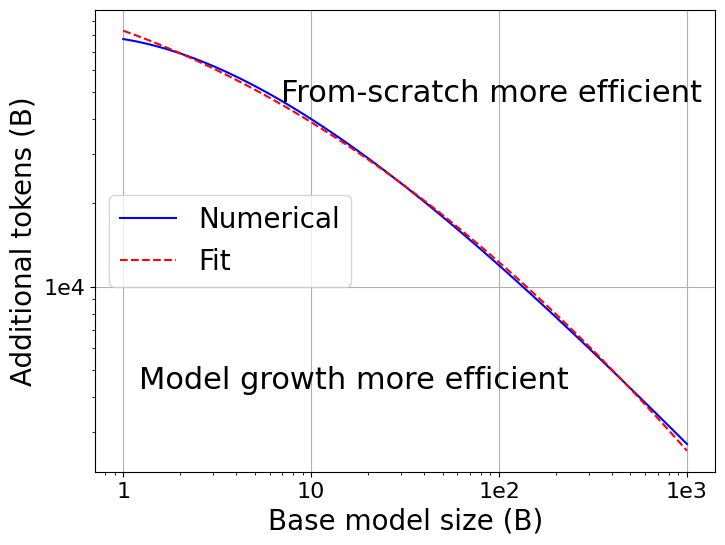}
    \caption{
        \textbf{Model growth efficiency decreases with sunk cost and model size.} For token budgets above the curve(s), training from scratch is more efficient than stacking-based model growth (growth factor 2); for budgets below, model growth remains advantageous.
        Shown are the numerical (blue) and analytical (red) solutions of Equation \ref{eq:sol}. Vertical axis is $D^*$ as in Equation  \ref{eq:sol}. \label{fig:sunk_cost_2}}
\end{figure*}

% Since the above equation involves non-integer polynomial exponents, we solve it numerically and approximate the solution analytically:
The right panel of Figure \ref{fig:sunk_cost_2} shows that $D^*$ decreases with increasing model size, with $D^*$ equal to 13T tokens for a 100B model.
When $D_{2} \lesssim D^*$, the required from-scratch tokens to catch up is more than 100\% of $D_1$: model growth remains more efficient than training from scratch.
% However, for $D_{2} \gtrsim D^*$, the efficiency reverses, favoring training from scratch.
% which indicates that model growth is beneficial when $D_2$ remains below a certain threshold, where the pretrained base model can still accelerate convergence. 
However, beyond this threshold, from-scratch training becomes more efficient.
Solving Equation \ref{eq:sol}, we can approximate the threshold $D^*$ analytically as $D^* \simeq  13\left(\frac{N}{10^{11}}\right)^{-0.6 - 0.04 \log (N/10^{11})}\; \text{T tokens}$.

\subsection{Compute Optimality}
\label{app:compute}

For scaling studies of LLMs, the cost of training is often estimated using floating point operations (FLOPs), which we simply refer to as compute.
The compute cost can be approximated as $C=6ND$ \citep{kaplan2020scaling}.
One is often interested in training LLMs with the least amount of compute.
For CPT, optimizing the compute based on Equation \ref{eq:multiplicative} leads to the scaling relations:
\begin{align}
    D^{\rm opt}_2 &\propto C_2^{\frac{\beta}{\beta +\alpha_{\rm eff}}},\; \;
    N^{\rm opt} \propto C_2^{\frac{ \alpha_{\rm eff}}{\beta + \alpha_{\rm eff}}}
    \label{eq:optimal_d_n}
\end{align}
 where $\alpha_{\rm eff}\coloneq\alpha_2-\alpha_3\log D_1$.
 Notably, as $D_1$ increases, $\alpha_{\rm eff}$ decreases, meaning larger models require more tokens for compute-optimal model reuse.

For model growth with growth factor 2, we want to scale $N_2,D_2$ optimally, $N^{\rm opt}_2,D^{\rm opt}_2$, given a FLOPs budget and fixing $D_1$, while minimizing the loss $L$, which we write as $L_{D_1}(D_2,N_1)$.
Here, $N_1$ is the model size before growth, and $N_2=2N_1$.
This is equivalent to solving the following:
\begin{align*}
    \left. \frac{\partial}{\partial D_2} L_{D_1}(D_2, C_2/12D_2) \right|_{D_2=D_2^{\rm opt}} = 0,\\
    \left. \frac{\partial}{\partial N_1} L_{D_1}( C_2/12N_1, N_1)\right|_{N_1=N^{\rm opt}_1} = 0
\end{align*}
where we have used $N_2=2N_1$ and $C_2 = 6 N_2 D_2$.
Solving the above equations leads to
\begin{align*}
    D^{\rm opt}_2 &= G \left(\frac{C_2}{12}\right)^{a}, \\
    N^{\rm opt}_1 &= G^{-1} \left(\frac{C_2}{12}\right)^{b}
\end{align*} 
where
\begin{align*}
    G &\coloneq \left( \frac{ A_{\rm eff}\alpha_{\rm eff}}{B\beta}\right)^{1/(\alpha_{\rm eff} + \beta)} \\
    a &\coloneq \frac{\beta}{\alpha_{\rm eff}+ \beta} \\
    b &\coloneq \frac{\alpha_{\rm eff}}{\alpha_{\rm eff}+ \beta}\\
    A_{\rm eff} &\coloneq A D_1^{-\alpha_1}\\
    \alpha_{\rm eff} &\coloneq \alpha_2-\alpha_3\log D_1
\end{align*}
We can henceforth relate $D^{\rm opt}_2$ and $N^{\rm opt}_1$ via 
$$
D^{\rm opt}_2 = G \left(GN^{\rm opt}_1 \right)^{a/b} \propto \left(N^{\rm opt}_1 \right)^{\beta/\alpha_2-\alpha_3\log D_1}
$$
and
$$
N^{\rm opt}_1 = G^{-1} \left(G^{-1}D^{\rm opt}_2 \right)^{b/a} \propto \left(D^{\rm opt}_2 \right)^{(\alpha_2-\alpha_3\log D_1)/\beta}
$$
 Notably, as $D_1$ increases, $\alpha_{\rm eff}$ decreases, meaning larger models require more tokens for compute-optimal model reuse.
% For model growth, one simply replaces $N$ with the new model size after growth, $N'$, in the above formulae to obtain similar relations.
We leave empirical verification of these relationships to future work, primarily due to cost considerations (see Appendix \ref{app:gpu} for an estimate of GPU hours spent in the work).
% As the loss landscape is sharper for overtrained models, a possible way to mitigate its negative effects is to use a lower learning rate (LR) for model reuse, such that the optimization steps do not overshoot the minima.
% We experiment with this idea in CPT and model growth with the setting described in Appendix \ref{app:cpt} and \ref{app:mg}, and fit the scaling laws.

% Indeed, the saturation effects are partially mitigated, 
% We have experimented with using a lower constant LR for CPT and stacking, and find that it yields worse performance (see Figure \ref{fig:cpt_lr} in Appendix \ref{app:mg}).

\subsection{Training with Extremely Large second-stage Token-to-Parameter Ratio}
\label{app:extreme}
\textbf{15M-to-30M model.}
We show the loss values of the last three $D_2$ points (each run with different $D_1$'s) in Table \ref{tab:error}.
To estimate the stochastic noise on the loss, for each $D_2$, we train again from the nearest stable-phase checkpoint but with three different random seeds.
Although there is slight indication of crossover (comparing $D_1=0.2884,9.2275$ at $D_2=9.2275,27.6824$), we are unable to make strong claims on it without a better understanding of the noise effects.

\begin{figure}
    \centering
        \includegraphics[width=0.45\linewidth]{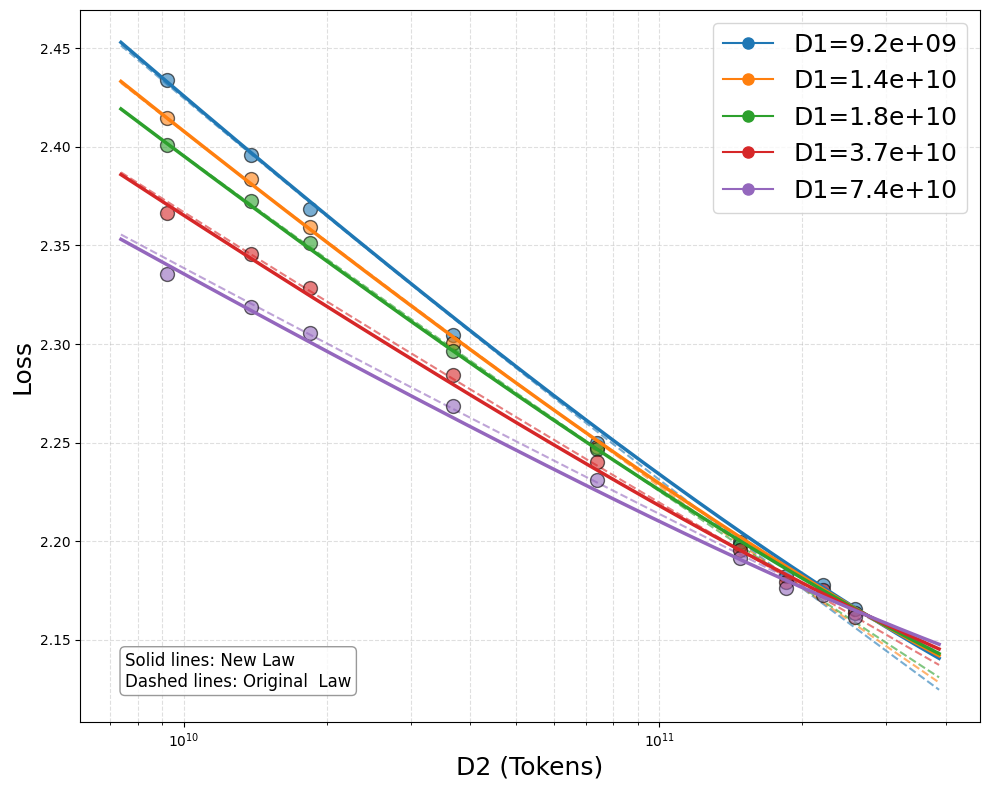}
    \caption{
        \textbf{Stacking with extremely large token-to-parameter ratio (0.5B-to-1B).} We compare the original scaling law and the new scaling law with a new fitting parameter $D_{\rm 2, max}$. \label{fig:1b_extreme}}
\end{figure}

\textbf{0.5B-to-1B model.}
We also train the larger 0.5B-to-1B model to more extreme token-to-parameter ratios (up to 200).
The result is shown in Figure \ref{fig:1b_extreme} (and Table \ref{tab:dense_05b_data} for the last four measured $D_2$), where we do not see significant crossover at the large $D_2$ limit.

Furthermore, at large $D_2$, the points are slightly curved upwards. 
This can best be seen by modifying the scaling law using a similar argument given in Section \ref{sec:limit}, by adding a new fitting parameter $D_{\rm 2, max}$:
\begin{equation}
    \hat{D_2} = \left(D_2^{-1} + D_{\rm 2, max}^{-1} \right)^{-1},
\end{equation}

The mean square error of the "tail" for the original and new scaling law are 2.7e-5 and 1.3e-5 (with similar global mean square errors) indicating that the new one fits the tail better (and from the Figure the new scaling law has an upward curve at large $D_2$).
This new formulation predicts a crossover and saturation of the curves afterwards (crossover occurs only when the fitted $D_{\rm 2,max}\lesssim e^{\alpha_2/\alpha_3}$).
The predicted crossover is beyond the computational reach and is in the region where noises are dominant.

\begin{table}[h]
\centering
\caption{15M-to-30M models' cross-entropy losses and the estimated error from multiple runs with different random seeds.}
\label{tab:error}
\begin{tabular}{cccc}
\toprule
\textbf{$D_2$ (B)} & \textbf{$D_1$ (B)} & \textbf{Loss} & \textbf{Error} \\ \midrule
\multirow{6}{*}{9.2275} & 0.2884 & 3.069959 & \multirow{6}{*}{$\pm$ 0.001345} \\
 & 0.4325 & 3.069911 &  \\
 & 0.5767 & 3.078497 &  \\
 & 1.1534 & 3.068825 &  \\
 & 2.3069 & 3.070050 &  \\
 & 9.2275 & 3.073884 &  \\ \midrule
\multirow{4}{*}{18.4549} & 0.2884 & 3.055554 & \multirow{4}{*}{$\pm$ 0.004026} \\
 & 1.1534 & 3.065518 &  \\ % New Entry
 & 2.3069 & 3.060454 &  \\
 & 9.2275 & 3.063905 &  \\ \midrule
\multirow{4}{*}{27.6824} & 0.2884 & 3.055849 & \multirow{4}{*}{$\pm$ 0.005699} \\
 & 1.1534 & 3.053557 &  \\
 & 2.3069 & 3.051366 &  \\
 & 9.2275 & 3.054023 &  \\ \bottomrule
\end{tabular}
\end{table}

\begin{table}[h]
\centering
\caption{0.5B-to-1B models' cross-entropy losses. The estimated error is 0.0012-0.0016.}
\label{tab:dense_05b_data}
\begin{tabular}{ccc}
\toprule
\textbf{$D_2$ (B)} & \textbf{$D_1$ (B)} & \textbf{Loss} \\ \midrule
\multirow{4}{*}{147.6395} & 9.2275 & 2.199658 \\
 & 18.4549 & 2.198132 \\
 & 36.9099 & 2.195471 \\
 & 73.8198 & 2.191318 \\ \midrule
\multirow{3}{*}{184.5494} & 9.2275 & 2.182314 \\
 & 36.9099 & 2.179196 \\
 & 73.8198 & 2.176350 \\ \midrule
\multirow{3}{*}{221.4593} & 9.2275 & 2.178030 \\
 & 36.9099 & 2.175260 \\
 & 73.8198 & 2.172848 \\ \midrule
 \multirow{3}{*}{258.3691} & 9.2275 & 2.165875 \\
 & 36.9099 & 2.163731 \\
 & 73.8198 & 2.161603 \\ \midrule
\multirow{3}{*}{258.3691} & 9.2275 & 2.165875 \\
 & 36.9099 & 2.163731 \\
 & 73.8198 & 2.161603 \\ \bottomrule
\end{tabular}
\end{table}

\subsection{More on Mechanistic Explanations}
\label{app:mechanistic}
\begin{figure*}[h]
    \centering
        \includegraphics[width=0.5\linewidth]{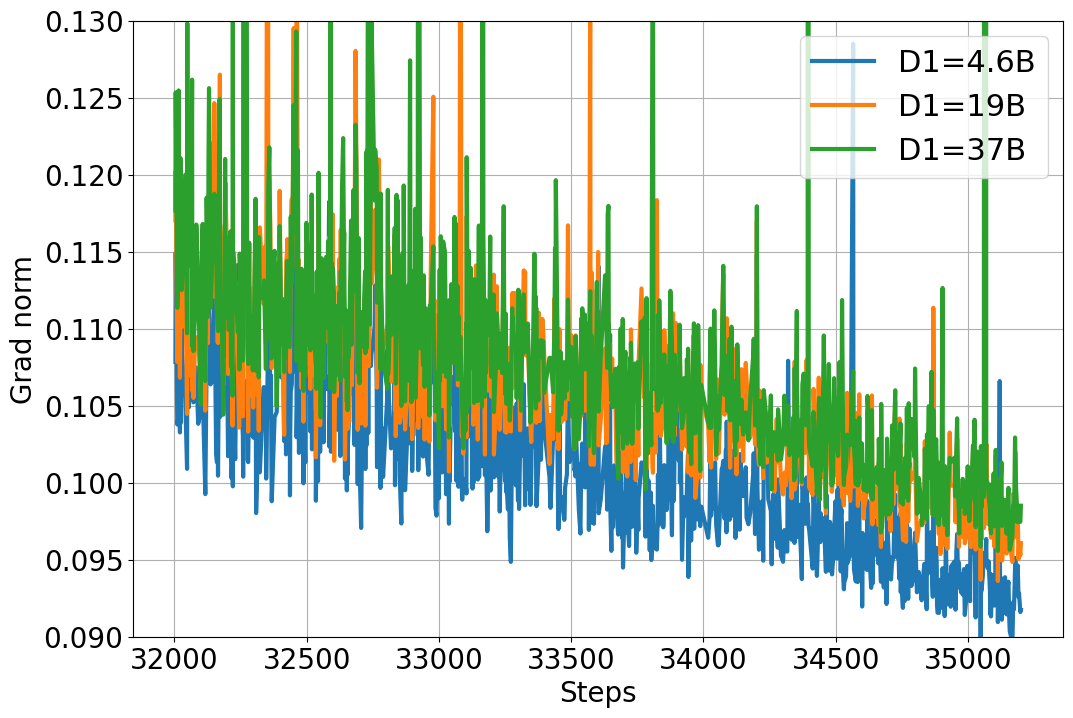}
    \caption{
        \textbf{Overtrained models have larger gradient norms when undergoing model reuse.} 
        We show the gradient norm curves for model growth (0.5B to 1B model via stacking) with respect to number of training steps in the second stage for base models trained with different values of first-stage tokens ($D_1$). \label{fig:grad_norm_stack}}
\end{figure*}

We show the gradient norms of model growth by stacking in Figure \ref{fig:grad_norm_stack}, where we also see that the gradient norms decrease with $D_1$ for fixed $D_2$, indicating saturation effects.

Next, we study other possible mechanistic explanations of the saturation effects.
Our scope of investigation is inspired by \citet{lyle2023understanding,lyle2024disentangling,klein2024plasticity}, which studied mechanistic signatures of plasticity loss in continual learning.

\begin{figure}
    \centering
        \includegraphics[width=0.45\linewidth]{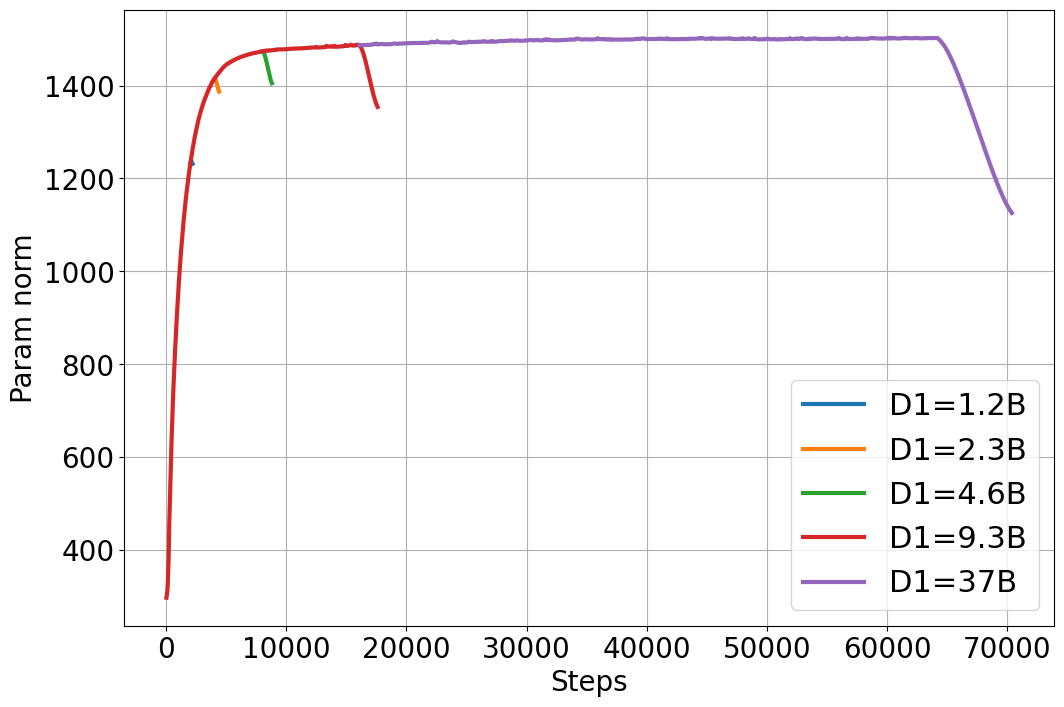}
    \caption{
        \textbf{Parameter norms do not show clear trend with $D_1$.} 
        We show the parameter norm curves training a 0.1B base model with respect to number of training steps, where the end of each line corresponds to the parameter norm of a base model trained with the respective number of training steps, $D_1$. The parameter norms of models (of various $D_1$) we use increase before decreasing with respect to $D_1$ (i.e., non-monotonic). There is therefore no direct correlation with $D_1$ (which correlates directly with saturation). \label{fig:param_norm_stack}}
\end{figure}

\paragraph{Parameter norms.}
Increased parameter norms (or weight norms) are indicated to be a signature of plasticity loss \cite{nikishin2022primacy}.
However, as in Figure \ref{fig:param_norm_stack}, we do not observe a clear trend of parameter norms with $D_1$ (which correlates directly with saturation) when training our models.
See the caption of Figure \ref{fig:param_norm_stack} for more details.

\begin{figure}
    \centering
        \includegraphics[width=0.45\linewidth]{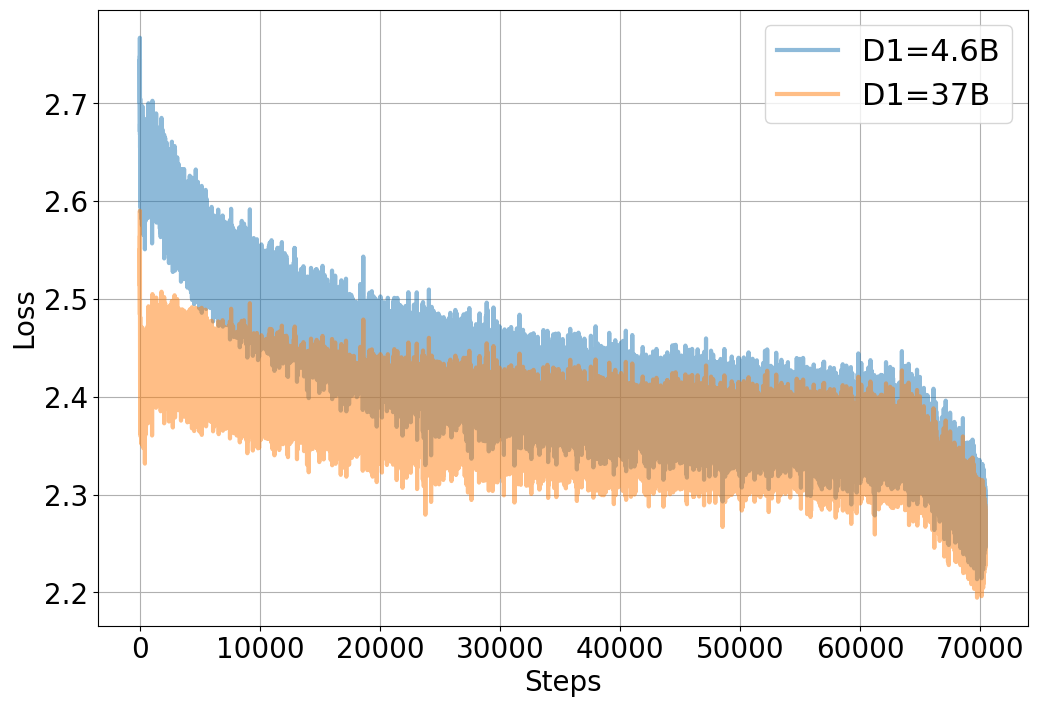}
    \caption{
        \textbf{Variance of training curves do not show clear trend with $D_1$.} 
        We show the variance of training curves for model growth by stacking (0.5B to 1B) with respect to number of training steps in the second stage for base models trained with different values of first-stage tokens ($D_1$). Up to 10,000 steps, the variance is slightly smaller for the training curve with lower $D_1$. However, there is no clear trend of variance in the training curves for different $D_1$ onwards. \label{fig:var_stack}}
\end{figure}

\paragraph{Sharpness via training trajectory.}
\citet{lyle2023understanding} suggested to look at the variance of training curves as a proxy for sharpness, i.e., increased variance in the training curves indicates that it is harder for the model to navigate the sharper loss landscape, indicating plasticity loss.
As in Figure \ref{fig:var_stack}, we do not observe a clear trend of difference in variance of training curves for different values of $D_1$ either (except in early training where training with smaller $D_1$ has slightly lower variance).
See the caption of Figure \ref{fig:var_stack} for more details.

\begin{figure}
    \centering
        \includegraphics[width=0.45\linewidth]{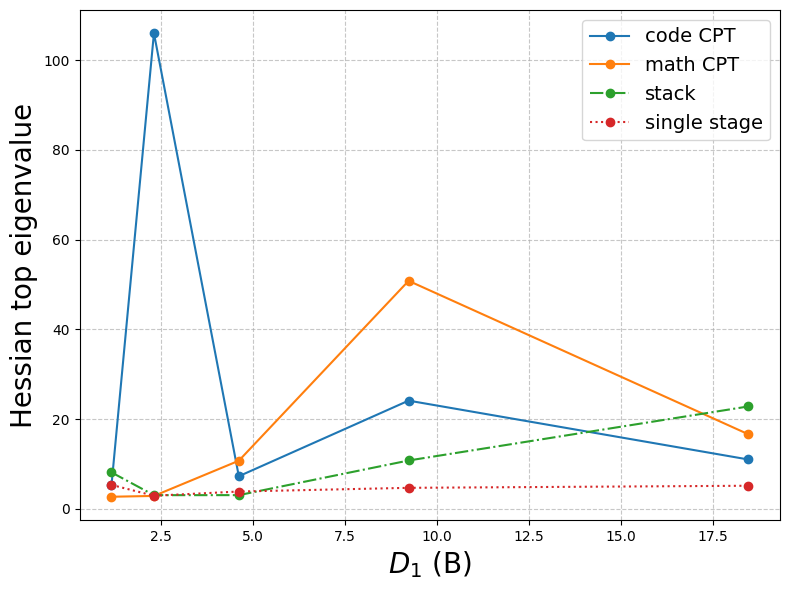}
    \caption{
        \textbf{Hessian with respect to $D_1$.} 
        We show the Hessian ($\lambda_{max}$) of a 0.1B model base models trained with different values of first-stage tokens ($D_1$), on model reuse scanarios (CPT on code, math; stacking) and single-stage training scenario. The number of power iteration in calculating $\lambda_{max}$ is 10, and the number of samples used to test each scenario is 128. \label{fig:hessian}}
\end{figure}

\paragraph{Sharpness via Hessian.}
Another way of studying the loss landscape is through the top eigenvalue of the Hessian ($\lambda_{max}$).
Here, we present a preliminary analysis using this quantity.
We use the Hessian-vector-product and power iteration to estimate $\lambda_{max}$ on training samples using base models trained with different $D_1$.
In Figure \ref{fig:hessian}, we show the estimated $\lambda_{max}$ with respect to $D_1$.

The overall trend is not very clear, but roughly speaking, we see that the model reuse scenarios (CPT on code, math; stacking) have some correlation with $D_1$, while training with the same data does not seem to correlate with $D_1$.

To summarize, we only observe a clear trend of gradient norms with $D_1$ among the mechanistic quantities studied.
We leave the studies of other possible mechanisms (and a more detailed study of the Hessian) to future work.

\subsection{More on the Toy Model}
\label{app:theory}
\textbf{Theoretical derivation of the exponential distribution w.r.t. depth.}
The exponential decay of probability with respect to depth $d$ is a derivation from the maximum entropy principle motivated by Mandelbrot's work on the distribution of word frequencies \citep{mandelbrot1953informational}. 

We seek a probability distribution $P(d)$ that maximizes Shannon Entropy $H = - \sum P(d) \ln P(d)$ subject to normalization $\sum P(d) = 1$ and a fixed average cost $\sum P(d) C(d) = \bar{C}$. Using Lagrange multipliers:
$$L = -\sum P(d) \ln P(d) - \lambda_0 \left( \sum P(d) - 1 \right) - \lambda \left( \sum P(d) C(d) - \bar{C} \right)$$
Setting $\frac{\partial L}{\partial P(d)} = 0$ solving for $P(d)$ yields the exponential (Boltzmann) distribution:$$P(d) \propto e^{-\lambda \cdot C(d)}$$In our tree model, depth $d$ represents the search cost $C$. This decay results in the power-law distribution of features.

\textbf{More on Hutter's learning curve theory.}
Let us describe the setup of \citet{hutter2021learning} in more detail: Each datum is associated with a feature $k$, and the feature is memorized when it is observed during training.
Assuming that i.i.d. data are drawn from a distribution $P(k)$, the expected loss is then proportional to the probability of encountering an unobserved feature during training after $D$ training samples:
$$L \propto \sum_{k=1}^{\infty} P(k) \left( 1-P(k) \right)^{D}$$
If $P(k)$ follows a power-law distribution $P(k) \sim k^{-(1+\alpha)}$, the loss can be shown to scale as $L \sim  D^{-\alpha/(1+\alpha)} \approx D^{-\alpha}$ for small $\alpha$.

Note that a simpler alternative assumption that leads to the same conclusion at small $\alpha$ is that features at the tail of the feature distribution is simply not observed during training. 
The loss is then an integral over the tail: 
$$L = \int_{D}^{\infty} k^{-(1+\alpha)} \textrm{d}k \sim D^{-\alpha},$$
leading to the same scaling behavior.

\end{document}